\documentclass[runningheads]{llncs}

\usepackage{eccv}

\newcommand{\ours}[0]{LoMa}
\usepackage{xparse}
\NewDocumentCommand{\loma}{o o}{%
  \IfNoValueTF{#1}{%
    LoMa%
  }{%
    Lo$_{#1}$Ma$_{#2}$%
  }%
}

\newcommand{\RN}[1]{\uppercase\expandafter{\romannumeral #1}}

\usepackage{eccvabbrv}

\usepackage{graphicx}
\usepackage{booktabs}

\usepackage[accsupp]{axessibility}  %

\usepackage{hyperref}
\usepackage{enumitem}
\usepackage{wrapfig}

\usepackage{orcidlink}

\usepackage{array}
\usepackage{graphicx}
\usepackage[percent]{overpic} %
\usepackage[table]{xcolor}

\begin{document}

\title{\ours: Local Feature Matching Revisited}

\authorrunning{D. Nordström and J. Edstedt et al.}

\institute{$^1$Chalmers University of Technology \quad $^2$Linköping University \quad $^3$University of Amsterdam \quad $^4$Centre for Mathematical Sciences, Lund University
}

\renewcommand{\thefootnote}{\fnsymbol{footnote}}

\author{
David Nordström$^1$\thanks{Equal contribution. Listing order is random.}
\and
Johan Edstedt$^2$\textsuperscript{\thefootnote}
\and
Georg Bökman$^3$ \\
Jonathan Astermark$^4$ 
\and
Anders Heyden$^4$
\and
Viktor Larsson$^4$
\\
 Mårten Wadenbäck$^2$
\and
Michael Felsberg$^2$
\and
Fredrik Kahl$^1$
 \\
}

\maketitle

\begin{abstract}
Local feature matching has long been a fundamental component of 3D vision systems such as Structure-from-Motion (SfM), yet progress has lagged behind the rapid advances of modern data-driven approaches. The newer approaches, such as feed-forward reconstruction models, have benefited extensively from scaling dataset sizes, whereas local feature matching models are still only trained on a few mid-sized datasets. In this paper, we revisit local feature matching from a data-driven perspective. In our approach, which we call \ours, we combine large and diverse data mixtures, modern training recipes, scaled model capacity, and scaled compute, resulting in remarkable gains in performance. Since current standard benchmarks mainly rely on collecting sparse views from successful 3D reconstructions, the evaluation of progress in feature matching has been limited to relatively easy image pairs. To address the resulting saturation of benchmarks, we collect 1000 highly challenging image pairs from internet data into a new dataset called HardMatch. Ground truth correspondences for HardMatch are obtained via manual annotation by the authors. In our extensive benchmarking suite, we find that \ours~makes outstanding progress across the board, outperforming the state-of-the-art method ALIKED+LightGlue by +18.6 mAA on HardMatch, +29.5 mAA on WxBS, +21.4 (1m, 10$^\circ$) on InLoc, +24.2 AUC on RUBIK, and +12.4 mAA on IMC 2022. We release our code and models publicly at \href{https://github.com/davnords/LoMa}{https://github.com/davnords/LoMa}.

\end{abstract}

\section{Introduction}
Structure-from-Motion (SfM)~\cite{hartley2003multiple}  aims to reconstruct the 3D world from unordered images and has long been a central problem in computer vision.
A crucial part of SfM pipelines, typically referred to as \emph{local feature matching}, is image matching through detection of sparse keypoints and description of their local appearance using high-dimensional representations, traditionally with \eg~SIFT~\cite{liu2010sift}, where correspondences are found by correlating the descriptions. %
To improve robustness and accuracy, neural network models have been introduced, both for detection and description, such as SuperPoint~\cite{detone2018superpoint}, ALIKED~\cite{Zhao2023ALIKED}, and DeDoDe~\cite{edstedt2024dedode}, and for sparse matching with models such as SuperGlue~\cite{sarlin2020superglue} and LightGlue~\cite{lindenberger2023lightglue}.
This paradigm yields fast and accurate matches and remains widely popular.
\begin{figure}[t]
    \centering

    \begin{subfigure}[c]{0.48\textwidth}
        \centering
        \includegraphics[width=\linewidth]{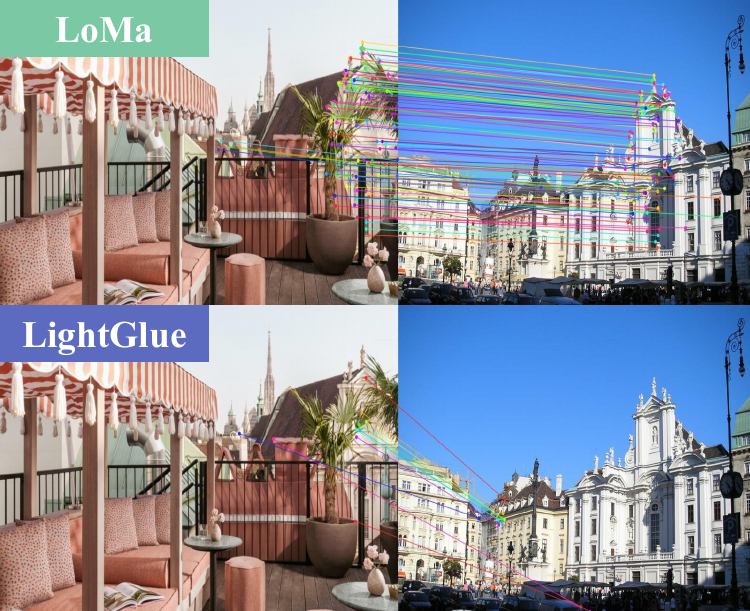}
        \caption{Matches on a pair from HardMatch.}
        \label{fig:teaser_a}
    \end{subfigure}
    \hfill
    \begin{subfigure}[c]{0.48\textwidth}
        \centering
        \includegraphics[width=\linewidth]{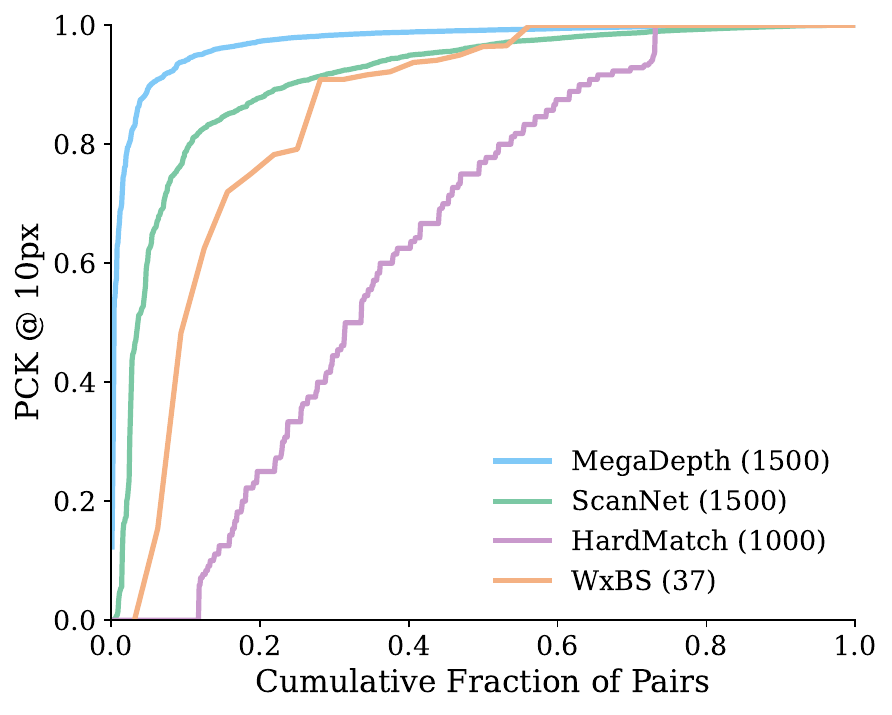}
        \caption{Distribution of pairs for different benchmarks successfully matched by SotA models.}
        \label{fig:teaser_b}
    \end{subfigure}

    \caption{
    \textbf{Revisiting local feature matching.} We introduce HardMatch, a challenging hand-annotated matching benchmark, and \ours, a fast and accurate family of local feature-based models. (a) \ours~successfully matches pairs from HardMatch where LightGlue fails, (b) HardMatch is significantly harder than previous benchmarks.}
    \label{fig:teaser}
\end{figure}
While still heavily used in practice, local feature matching has recently been overshadowed in the literature by the advent of detector-free methods such as LoFTR~\cite{sun2021loftr} and RoMa~\cite{edstedt2024roma}, and feed-forward reconstruction models such as MASt3R~\cite{leroy2024grounding} and VGGT~\cite{wang2025vggt} that are typically trained on orders of magnitude more data than their local feature matching counterparts.
In the context of detector-free methods, it is often argued that detector-based local feature matching is fundamentally limited~\cite{sun2021loftr}, and a significant amount of research has gone into how to scale detector-free SfM~\cite{he2024detector,lee2025dense,duisterhof2025mast3r,elflein2025light3r}, in order to overcome these supposed limitations.
We argue that \emph{the reports of the death of the local feature matcher are greatly exaggerated}. %

In this paper, we revisit local feature matching from a data-driven perspective.
In particular, we focus on (i) curating a large and diverse training data mixture together with scalable training recipes for both descriptors and matchers,
and (ii) increasing training compute along two axes: data scale (the number and diversity of image pairs) and model capacity (the number of parameters). As we demonstrate through extensive experiments and ablations, these changes lead to substantial improvements in matching performance across a wide range of benchmarks. Our models outperform prior local feature methods by large margins and, in several settings, are competitive with or even surpass recent dense matching and feed-forward reconstruction pipelines. \Cref{fig:teaser_a} 
shows a qualitative example of a very challenging case that our matcher solves.

To meaningfully assess progress in matching capabilities and guide future research, well-designed evaluations and benchmarks are essential. Historically, improvements in feature matching have been measured on datasets derived from SfM reconstructions, such as MegaDepth~\cite{li2018megadepth}. However, as we show in \cref{fig:teaser_b}, many of these benchmarks are now close to saturation: for a large fraction of image pairs, modern state-of-the-art matchers already recover a high percentage of correct correspondences. When benchmarks saturate, further improvements become difficult to observe, even when models meaningfully improve in robustness or generalization. This obscures remaining failure modes and risks encouraging overfitting to benchmark-specific artifacts, such as particular geometric verification settings, rather than advancing fundamental matching capability. To clearly measure progress, more challenging and diverse benchmarks are required.
However, existing difficult image matching benchmarks, such as WxBS~\cite{mishkin2015WXBS}, are too small to reliably measure model improvements.

To address these limitations, we manually annotate image correspondences for a collection of 1000 pairs from 100 different categories, which we call \textbf{HardMatch}. The dataset is organized into 9 challenging groups
spanning diverse and extreme matching scenarios.
We find that feed-forward reconstruction methods largely fail on this benchmark, and even SotA dense matchers struggle.
In a \textit{return of the local feature matcher}, we demonstrate that our family of models, \textbf{\ours}, can achieve performance even surpassing dense methods (and greatly outperforming sparse methods) by training on more diverse data with modern training recipes and increased compute.
Our models, \ours-\{B(ase), L(arge), G(igantic)\}, set a strong baseline for future progress in feature matching.

\noindent\textbf{Our main contributions can be summarized as:}
\begin{enumerate}[topsep=0pt]
    \item We revisit local feature matching from a modern, data-driven perspective (\cref{sec:method}), introducing new training datasets with MVS-generated ground-truth and training recipes that we will make publicly available.
    \item We introduce \textbf{HardMatch}, a challenging benchmark of 1000 hand-labeled image pairs that is lightweight yet large and difficult enough to provide meaningful signal for future research. We additionally report a human baseline based on independent annotators (\cref{sec:hardmatch}).
    \item We release a fast and accurate family of descriptor-matcher models that achieve SotA performance on HardMatch (+18.6mAA over LightGlue)
    and strong results across more than ten established matching and visual localization benchmarks. Extensive evaluations and ablations are provided in \cref{sec:experiments}.
\end{enumerate}
Subsequent to the present work, we extended LoMa to make it invariant to in-plane image rotations~\cite{nordstrom2026whohandlesorientation}.

\section{Related Work}

\subsubsection{Feature Matching. } Finding pixel correspondences between two images is a fundamental task in 3D computer vision. Traditionally, image matching has been done in three stages: (i) keypoint detection, (ii) local feature description, and (iii) nearest neighbor matching in feature space. Learning-based approaches for keypoint detection~\cite{barroso2019key, mishkin2018repeatability, verdie2015tilde, edstedt2024dedodev2, edstedt2025dad}, description~\cite{balntas2017hpatches, tian2019sosnet, germain2020s2dnet, edstedt2024dedode, bökman2024steerers}, as well as joint detection and description~\cite{detone2018superpoint, revaud2019r2d2, tyszkiewicz2020disk, Wang_2021_ICCV, zhao2022alike, Zhao2023ALIKED}, have been proposed to replace handcrafted methods such as SIFT~\cite{lowe2004distinctive} and ORB~\cite{rublee2011orb}. SuperGlue~\cite{sarlin2020superglue} proposed replacing the nearest neighbor matcher with a graph attention network, allowing global reasoning on local keypoint descriptors. Subsequently, LightGlue~\cite{lindenberger2023lightglue} introduced a layer-wise loss and improved speed through pruning and early-stopping. Detector-free matching, first introduced in LoFTR~\cite{sun2021loftr}, in contrast to sparse matching, eliminates keypoints. Matching benchmarks~\cite{dai2017scannet, li2018megadepth, mishkin2015WXBS} have, since DKM~\cite{edstedt2023dkm}, been topped by dense matchers~\cite{edstedt2024roma, zhang2025ufm, edstedt2026romav2}, which match every pixel. 
Learning-based SfM methods, commonly referred to as \textit{feed-forward reconstruction}~\cite{wang2024dust3r, dens3r, wang2025vggt, wang2026vggtomega, leroy2024grounding, keetha2025mapanything}, often include matching objectives. 
Notably, VGGT~\cite{wang2025vggt} uses a tracking head and MASt3R~\cite{leroy2024grounding} combines pointmap regression with detector-free local features.
In this work, our contribution is not the development of a novel matcher or descriptor.
Instead, we use the existing DeDoDe descriptor,  DaD~\cite{edstedt2025dad} keypoints, and LightGlue matcher, with our proposed modern training recipe and our large-scale curated datasets. We show that using our approach, we can greatly surpass the performance of the original models.

\subsubsection{Matching Evaluation.} Feature matchers are commonly evaluated through relative pose estimation on sparse views from successful 3D reconstructions, such as MegaDepth~\cite{li2018megadepth} and ScanNet~\cite{dai2017scannet}, or visual localization~\cite{taira2018inloc, sattler2018benchmarking, Jafarzadeh_2021_ICCV, arnold2022map}. 
These methods generally use pre-existing 3D reconstructions with localized query images to construct the benchmark.
While this enables directly evaluating the estimated pose, the requirement for successful localization means that matching the query images is already solvable with existing systems.
Thus, both categories are mostly saturated.
In a similar vein, the Image Matching Challenge (IMC)~\cite{image-matching-challenge-2022} is a yearly challenge that aims to test the limits of reconstruction methods, with a hidden test set of ground-truth (GT) reconstructions.
While IMC challenges are typically less saturated, evaluating matchers on them is typically a complex task, as there is no standardization, any reconstruction method is allowed, and SfM-pipelines involve a large number of hyperparameters.

In contrast, some previous benchmarks forgo mapping, and instead evaluate using only GT correspondences.
One such work is WxBS~\cite{mishkin2015WXBS}, which consists of manually collected and labeled challenging pairs.
Instead of evaluating the error in estimated relative pose, they use the epipolar error of the GT correspondences under a Fundamental matrix, which is estimated using correspondences from the matcher. However, its small size (37 pairs) limits its usefulness in model comparison. Besides this, we also find signs of this benchmark being near saturation (\cf \Cref{tab:relpose,fig:wxbs-curve}). Our work takes a similar approach to WxBS, but includes more than 25 times as many pairs, with much higher diversity and difficulty.

\section{Training the LoMa Descriptor and Matcher}
\label{sec:method}
\begin{figure}[t]
\centering
    \includegraphics[width=\linewidth]{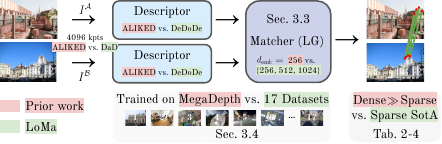}

\caption{\textbf{The \ours~pipeline.} By replacing ALIKED~\cite{Zhao2023ALIKED} with DaD~\cite{edstedt2025dad}+DeDoDe~\cite{edstedt2024dedode} and training the descriptor and matcher on a large collection of datasets we achieve SotA results, even surpassing dense matchers on some tasks (\eg HardMatch).}
\label{fig:method-figure}
\end{figure}

In this section, we detail the training of the LoMa descriptor and matcher (\cf \cref{fig:method-figure}), based on DeDoDe~\cite{edstedt2024dedode} and LightGlue~\cite{lindenberger2023lightglue}, respectively. 

\subsection{Problem Formulation}\label{subsec:problem}

The aim in two-view matching is to obtain correct keypoint correspondences between two images $I^{\mathcal{A}}$ and $I^\mathcal{B}$.
We follow a common three-stage approach, where first keypoints $x_i^{\mathcal{A}}$ and $x_j^{\mathcal{B}}$ are detected in the images, second the keypoints are assigned descriptions $f_i^\mathcal{A}$ and $f_j^\mathcal{B}$, and third the descriptions are matched between the two images. The keypoints are assigned descriptions using a neural network called the \emph{descriptor} $g_\theta$, after which the descriptions are matched using a second neural network called the \emph{matcher} $h_\phi$.

\subsection{Learning Objective}\label{subsec:learning}
In this work, we do not train a detector. Instead, we use DaD to supervise both the descriptor ($g_\theta$) and matcher ($h_\phi$). We compare DaD to other detectors and an ensemble in \cref{tab:changing-detector} in the supplementary. We first train $g_\theta$, followed by $h_\phi$ with $g_\theta$ frozen. The number of keypoints in each image during training is $N=2048$.

Ground truth (GT) correspondences are obtained via known relative poses and depth maps in the training datasets.
We denote the GT matches by 
$
\mathcal{M}^{\mathcal{A,B}}=\{(i, j)~|~x_i^\mathcal{A}~\text{and}~x_j^\mathcal{B}~\text{match}\}.
$

\subsubsection{Description.} For training the descriptor $g_\theta$, we follow DeDoDe~\cite{edstedt2024dedode} and use a dual-softmax based loss.
Descriptions of each keypoint are obtained as $f_i^\mathcal{A}=g_\theta(x_i^\mathcal{A},I^\mathcal{A}),f_j^\mathcal{B}=g_\theta(x_j^\mathcal{B},I^\mathcal{B})\in\mathbb{R}^{d_\text{desc}}$, and
a description similarity matrix is defined per image pair as
\begin{equation}
    S_{ij}={f_i^\mathcal{A}}^\top f_j^\mathcal{B}.
\end{equation}
The loss per image pair is given by
\begin{equation}\label{eq:desc_loss}
    \mathcal{L}_\text{desc} = -\left(\sum_{(i, j)\in\mathcal{M}^{\mathcal{A,B}}}\mathrm{log}\,\operatorname{softmax}_i(\tau^{-1} S_{ij}) + \mathrm{log}\,\mathrm{softmax}_j(\tau^{-1}S_{ij})\right),
\end{equation}
where $\tau^{-1}$ is the inverse temperature, a hyperparameter.
The loss encourages the dual-softmax matrix (also called soft assignment matrix) 
\begin{equation}\label{eq:assignment}
    \mathcal{P}_{ij}=\mathrm{softmax}_i(\tau^{-1} S_{ij})\odot \mathrm{softmax}_j(\tau^{-1} S_{ij})
\end{equation}
to have the GT matches as maxima along both $i$ and $j$.
As noted in \cite{leroy2024grounding}, \eqref{eq:desc_loss} can be viewed as a form of infoNCE-loss applied over the GT correspondences.

\subsubsection{Matching.}
For training the matcher, we follow LightGlue~\cite{lindenberger2023lightglue}.
The matcher $h_\phi$ takes keypoints and descriptions for two images as input and outputs refined descriptions $\tilde f$ for each keypoint, which now depend on both images:%
\begin{equation}
    \left(
        \left(\tilde f_i^\mathcal{A}\right)_{i=1}^N, \left(\tilde f_j^\mathcal{B}\right)_{j=1}^N 
    \right)
    =
    h_\phi \left(
        \left(x_i^\mathcal{A}, f_i^\mathcal{A}\right)_{i=1}^N, \left(x_j^\mathcal{B}, f_j^\mathcal{B}\right)_{j=1}^N
    \right).
\end{equation}
In each layer of $h_\phi$, we use the dual-softmax loss \eqref{eq:desc_loss} on the refined features, passed through a linear head, along with a separate matchability loss. A separate linear head with softmax activation predicts a matchability score for each keypoint. We supervise this prediction using a binary cross-entropy loss with the ground-truth matchability. A keypoint is defined as matchable if it has a match in the ground truth set $\mathcal{M^{A,B}}$. Layer-wise supervision allows trading performance for speed at inference time. We study this trade-off in \cref{subsec:inference}.

\subsection{Architecture}\label{subsec:models} %

Our descriptor follows the DeDoDe~\cite{edstedt2024dedode} architecture, while the matcher is based on LightGlue~\cite{lindenberger2023lightglue}. Input keypoints and descriptors are processed through $L$ identical blocks of self- and cross-attention, progressively refining the descriptors. When the descriptor dimension ($d_{\text{desc}}$) differs from the matcher embedding dimension ($d_{\text{emb}}$), we apply a learned linear projection.

Self-attention is applied by each point attending to all points of the same image, while in cross-attention each point attends to all points of the other image. Rotary position embeddings (RoPE)~\cite{rope} are used in the self-attention computation, making the attention scores dependent on the relative positions $x_{i} - x_{i'}$. Positional embeddings are not used in the cross-attention computation.

At inference, we use the descriptions output from the last layer to define a dual-softmax matrix $\mathcal{P}_{ij}$ as in \eqref{eq:assignment}. 
A correspondence $(i,j)$ is registered when $\mathcal{P}_{ij}$ represents a maximum along both the rows and columns, \ie the match is mutual. We discard matches for which $\mathcal{P}_{ij}<\mu$ with $\mu=0.1$.

We release three main variants of the \ours~matcher, B, L, and G, with progressively increasing size. All variants share the same architecture, consisting of $L=9$ transformer blocks that alternate between self-attention and cross-attention layers, and use attention heads of dimension 64 throughout. They differ only in their embedding dimensionality, which is 256, 512, and 1024, respectively. We also release B$^{128}$, using the lighter descriptor DeDoDe-B, instead of G, with $d_{\text{desc}}=128$, providing a lightweight set of features for \eg visual localization.

\subsection{Training Data}\label{subsec:training_data}
Our data mixture, presented in \cref{tab:dataset_mix}, is inspired by RoMa v2~\cite{edstedt2026romav2} and UFM~\cite{zhang2025ufm} by incorporating both wide baseline and optical flow datasets. Compared to prior matchers such as LightGlue, which was pretrained on synthetic homographies and fine-tuned on MegaDepth, our training data is significantly more diverse. In addition to the datasets used in RoMa v2, we add Aria Synthetic Environments~\cite{AriaSynthEnv:2025}, CO3Dv2~\cite{co3d:2021}, MPSD~\cite{mpsd:2020}, MegaDepth (Re-MVS), MegaScenes~\cite{tung2024megascenes}, MegaSynth~\cite{Jiang_2025_CVPR}, and SpatialVID~\cite{wang2025spatialvid}. The large data collection leads to a near 10-point improvement on HardMatch (\cf \cref{tab:pipeline-ablation}). For three of these datasets we compute 3D ground truth beyond the original data. We will make the data and code for these datasets, which we provide further details on below, public.

\subsubsection{MegaDepth (Re-MVS):} We run COLMAP~\cite{schoenberger2016sfm} MVS (photometric+geometric, default settings) on all scenes, additionally including reconstructions skipped in MegaDepth (all sparse models beyond the first reconstruction).

\subsubsection{MegaScenes:} MegaScenes contains a large number of scenes. However, we find that many of these are not of sufficient quality or size to constitute good training data. 
We select a subset of reconstructions and from these filter out a total of $303$ scenes with a sufficiently large number of cameras and 3D points. For these scenes we run standard COLMAP MVS, similarly as for MegaDepth (Re-MVS).

\subsubsection{SpatialVID:} While SpatialVID provides 3D annotations, we find them to be insufficiently accurate for feature matching. We therefore select a subset comprising 59 scenes, and run COLMAP SfM (using SIFT+DaD keypoints with RoMa v2 correspondences) with shared intrinsics. 
As most scenes contain dominant forward motion, we do not filter initial pairs on forward motion, as this commonly led to reconstructions failing.
We implement a custom MVS pipeline using RoMa v2 correspondences with a simple native PyTorch~\cite{paszke2019pytorch} PatchMatch implementation to compute depth maps.

\begin{table}[ht] \centering
\scriptsize
\caption{\textbf{Training data}. Unlike previous local feature matching methods~\cite{tyszkiewicz2020disk,edstedt2024dedode} typically trained on MegaDepth~\cite{li2018megadepth}, we scale our training to 17 3D datasets, approaching the data volume used in feed-forward reconstruction.}
\label{tab:dataset_mix}
\setlength{\tabcolsep}{2pt}
\begin{tabular}{lcc}
\hline
\toprule
Datasets                          & Type / GT Source            & Weight \\ 
\midrule
ScanNet++ v2~\cite{yeshwanth2023scannet++} & Indoor / Mesh & 1 \\
BlendedMVS~\cite{yao2020blendedmvs}  & Aerial / Mesh      &1 \\
Map-Free~\cite{arnold2022map}& Object-centric / MVS &1 \\
Hypersim~\cite{roberts2021hypersim}& Indoor / Graphics          &1 \\
MegaScenes~\cite{tung2024megascenes} & Outdoor / MVS & 1 \\
MegaDepth~\cite{li2018megadepth}    & Outdoor / MVS        &1 \\
MegaDepth (Re-MVS) & Outdoor / MVS        &1 \\
AerialMD~\cite{vuong2025aerialmegadepth}    & Aerial / MVS        &1 \\
TartanAir v2~\cite{wang2020tartanair}& Outdoor / Graphics          &1 \\

Mapillary Planet-scale Depth~\cite{mpsd:2020} & Driving / MVS  & 0.1 \\
Aria Synthetic Environments~\cite{AriaSynthEnv:2025} & Indoor / Graphics & 0.1\\
CO3Dv2~\cite{co3d:2021} & Object-centric / MVS & 0.1 \\
MegaSynth~\cite{Jiang_2025_CVPR} & Indoor / Graphics & 0.1\\
SpatialVID~\cite{wang2025spatialvid} & Forward-motion / MVS & 0.01\\
\midrule
FlyingThings3D~\cite{mayer2016large} & Outdoor / Graphics & 0.5\phantom{0}\\
UnrealStereo4k~\cite{tosi2021unrealstereo4k} & Outdoor / Graphics & 0.01\\
Virtual KITTI 2~\cite{gaidon2016virtual,cabon2020vkitti2} & Outdoor / Graphics & 0.01\\

\bottomrule
\end{tabular}
\vspace{-0.5em}
\normalsize
\end{table}

\subsection{Training}\label{subsec:training}

For all training, we use the AdamW~\cite{loshchilov2018decoupled} optimizer, the data mix outlined in~\cref{tab:dataset_mix}, and a fixed resolution of $560\times 560$. We use a cosine annealing learning rate with a peak learning rate of $2\times10^{-4}$ and a global batch size of 64. We use a slight weight decay of $5\times10^{-5}$ and use Exponential Moving
Average (EMA) with a decay factor of $\alpha=0.999$. We train the descriptor for 50K steps, which takes approximately one day on $8\times \text{A100:40GB}$. We show in the supplementary (\cf \cref{fig:descriptor-scaling}) that training the descriptor for longer does not help. We train the matcher for 250K steps. Sizes B and L are trained on $8\times \text{A100:40GB}$ while G is trained on $16\times \text{A100:40GB}$, each taking around two days. %

\section{HardMatch} \label{sec:hardmatch}

In this section, we introduce HardMatch, an extremely challenging image matching benchmark divided into 9 groups (\cf \cref{fig:hardmatch}). We detail data collection (\cref{subsec:data-collection}), evaluation (\cref{subsec:method-hardmatch-evaluation}), and qualitative characteristics (\cref{subsec:method-hardmatch-qual}). We will release the benchmark publicly under a permissive license.

\subsection{Data Collection}\label{subsec:data-collection}
The main steps in the data collection process are: (i) identifying candidate images online, (ii) selecting difficult pairs using a matching model with manual annotation, and (iii) manual keypoint annotation. We detail the steps below.

\subsubsection{Finding Images.} We begin by identifying a large corpus of candidate images by scraping 100 categories of Wikimedia Commons under permissive licenses. The categories were chosen to provide high diversity. We illustrate categories in the test set in \cref{fig:categories-distribution}. The methodology is inspired by MegaScenes~\cite{tung2024megascenes}.

\subsubsection{Identifying Difficult Pairs.} To filter the large image collection into difficult matching pairs, we randomly sample 100 pairs per scene and use the confidence map of RoMa~v2~\cite{edstedt2026romav2} to identify difficult pairs. We select pairs where RoMa~v2 is uncertain by thresholding the maximum confidence between 0.3 and 0.9, which provides a good balance of difficult pairs while still having some overlap. We manually inspect each identified pair and classify it as matchable or unmatchable, proceeding until we identify 10 pairs per category. The categories are randomly split into a validation set (10 categories) and a test set (90 categories).

\subsubsection{Annotating Keypoints.} We manually annotate corresponding keypoints in each pair, identifying as many salient matches as possible. Each pair contains between 8 and 28 annotated correspondences. The resulting dataset, which we call HardMatch, consists of 1000 image pairs from all over the world. We illustrate the geographic and temporal distribution in~\cref{fig:hardmatch-statistics} in the supplementary. To provide more granular insights, we group pairs into the labels shown qualitatively in \cref{fig:qual}. The smallest group is roughly equal to WxBS in size.

To verify our keypoint annotations, we provide a human baseline and estimate the ground truth error using independent annotators. Eight independent annotators are each assigned 20 pairs to verify. Annotators are asked to match a random keypoint in the first image with an arbitrary pixel location in the second image. We record the pixel error distribution and include a curve in \cref{fig:hardmatch-curve}.

\subsection{Evaluation}\label{subsec:method-hardmatch-evaluation}

Following WxBS, we evaluate by estimating a Fundamental matrix $F$ using correspondences from the matcher and computing the epipolar error of the GT correspondences. We compute the percentage correct keypoints (PCK) under different pixel error thresholds. Each pair contributes equally to the PCK, regardless of number of keypoints. More details on the evaluation can be found in the supplementary (\cref{appendix:hardmatch-evaluation}) as well as an alternative evaluation methodology directly using the GT keypoints (\cref{appendix:correspondence-evaluation}).

\subsection{Qualitative Characteristics}\label{subsec:method-hardmatch-qual}
HardMatch is an image matching dataset specifically designed to capture extreme appearance variations. Many pairs consist of images taken under substantially different conditions. The dataset includes examples such as aerial versus ground views, images captured over a century apart, hand-drawn sketches paired with natural photographs, night–day transitions, seasonal changes, and viewpoint differences of up to 180 degrees. Selected examples illustrating this diversity are shown in~\cref{fig:qual}. More pairs are found in the supplementary.

\begin{figure}[t]
    \centering

    \begin{subfigure}[t]{0.6\linewidth}
        \vspace{0pt}
        \centering
        \includegraphics[width=0.32\linewidth]{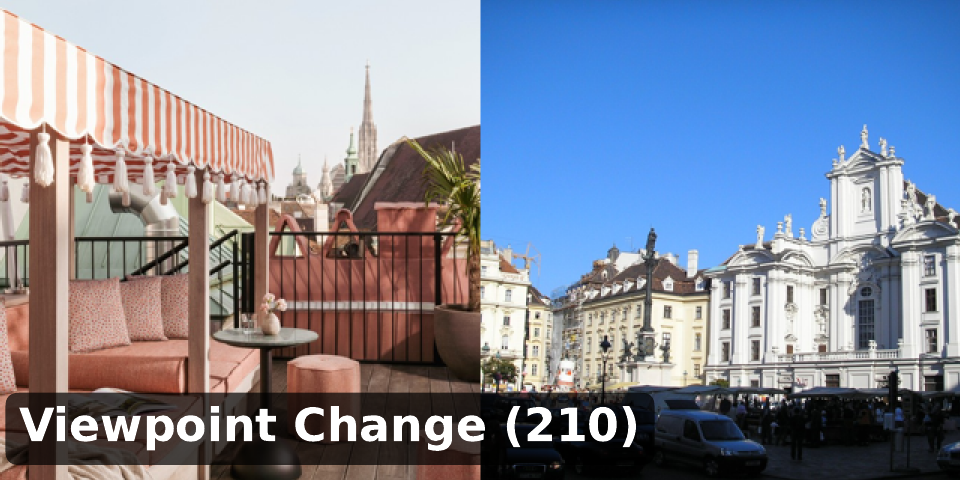}
        \includegraphics[width=0.32\linewidth]{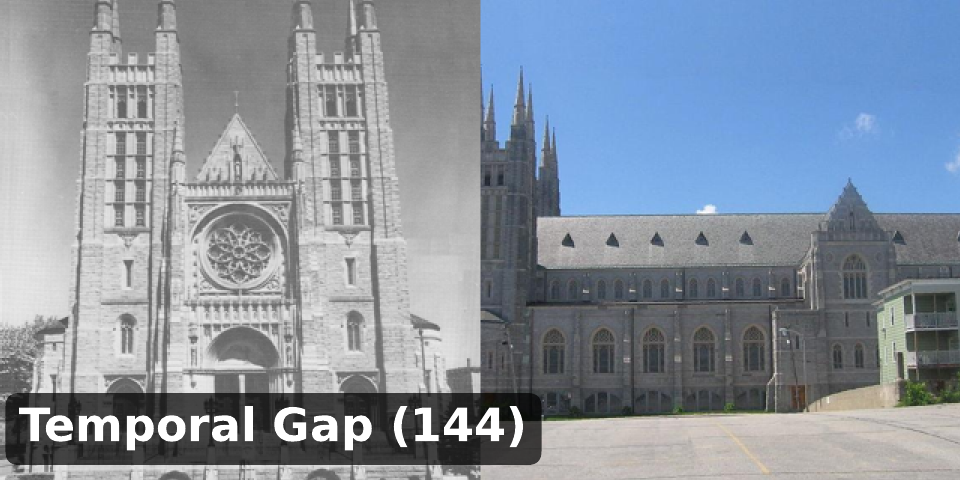}
        \includegraphics[width=0.32\linewidth]{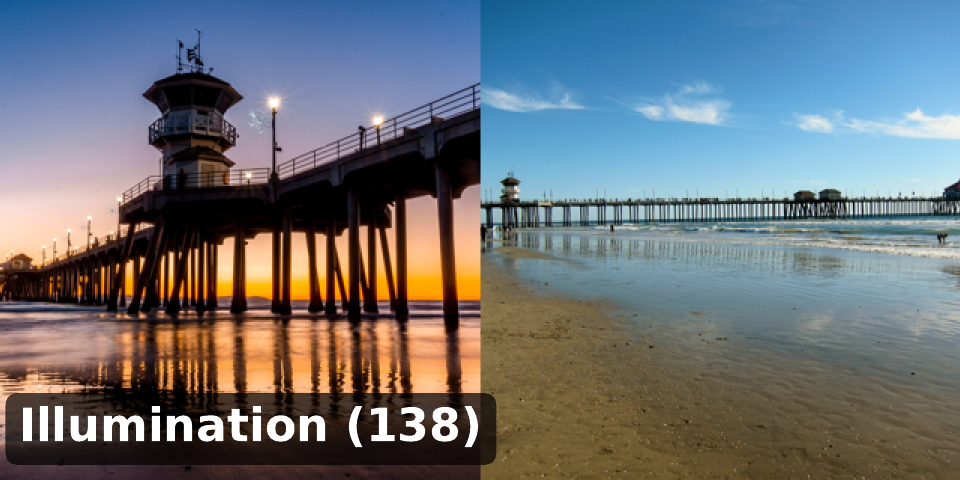}

        \includegraphics[width=0.32\linewidth]{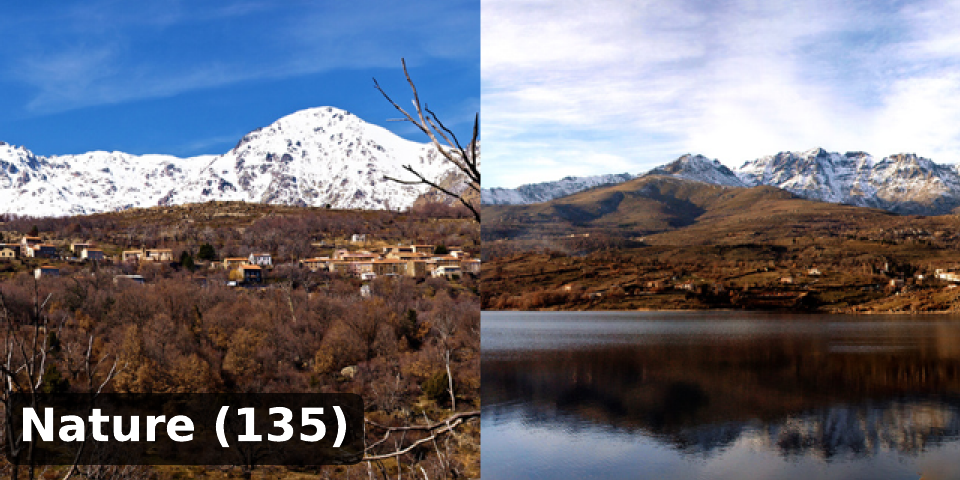}
        \includegraphics[width=0.32\linewidth]{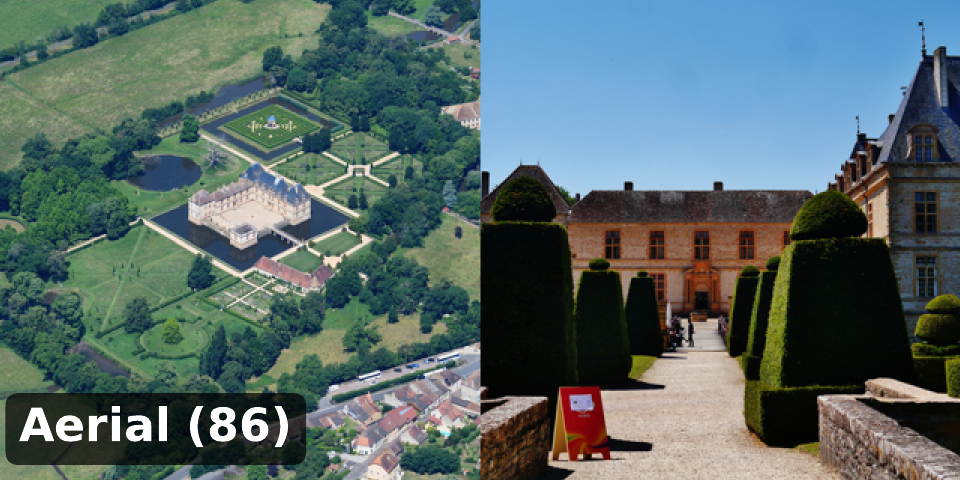}
        \includegraphics[width=0.32\linewidth]{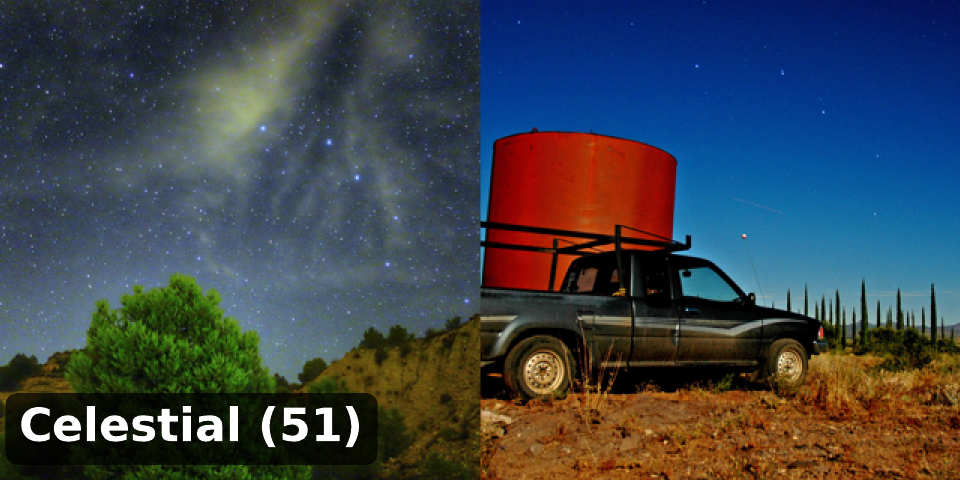}

        \includegraphics[width=0.32\linewidth]{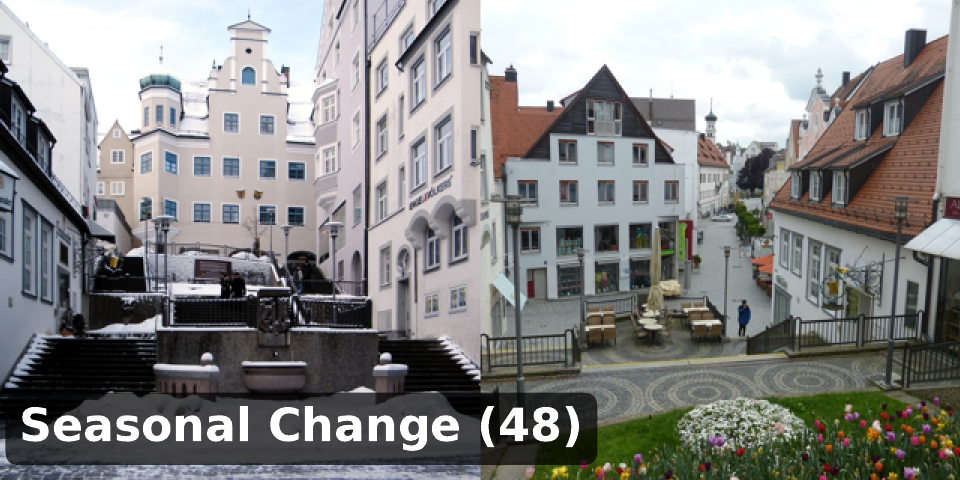}
        \includegraphics[width=0.32\linewidth]{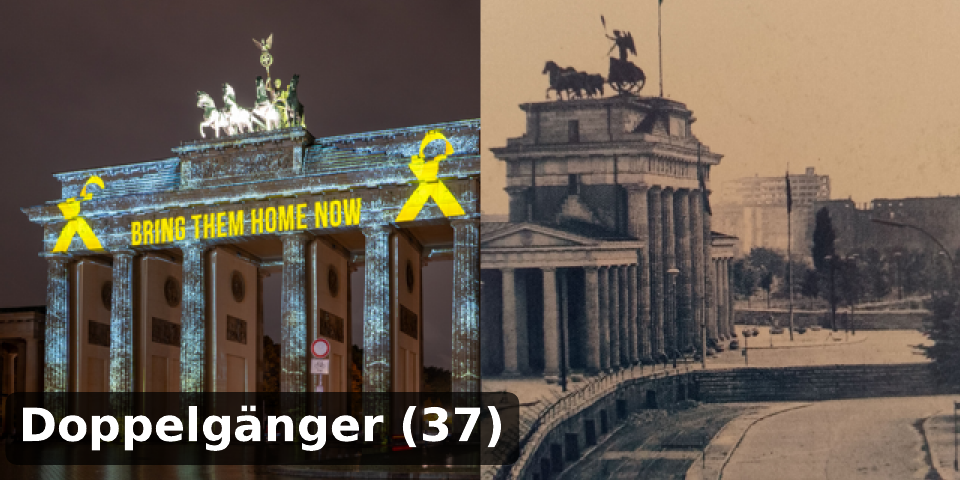}
        \includegraphics[width=0.32\linewidth]{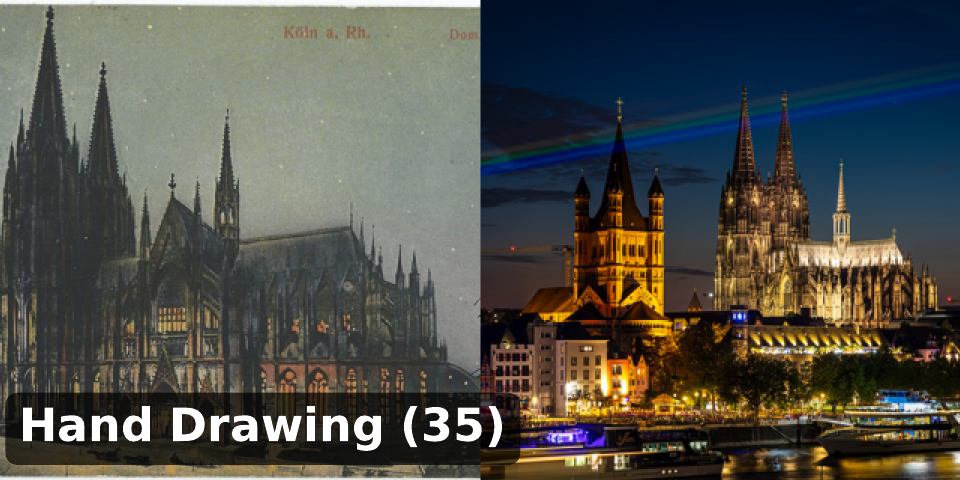}
        \caption{Example pairs per group (\#pairs in parentheses).}
        \label{fig:qual}
    \end{subfigure}
    \hfill
    \begin{subfigure}[t]{0.39\linewidth}
        \vspace{0pt}
        \centering
        \includegraphics[width=\linewidth]{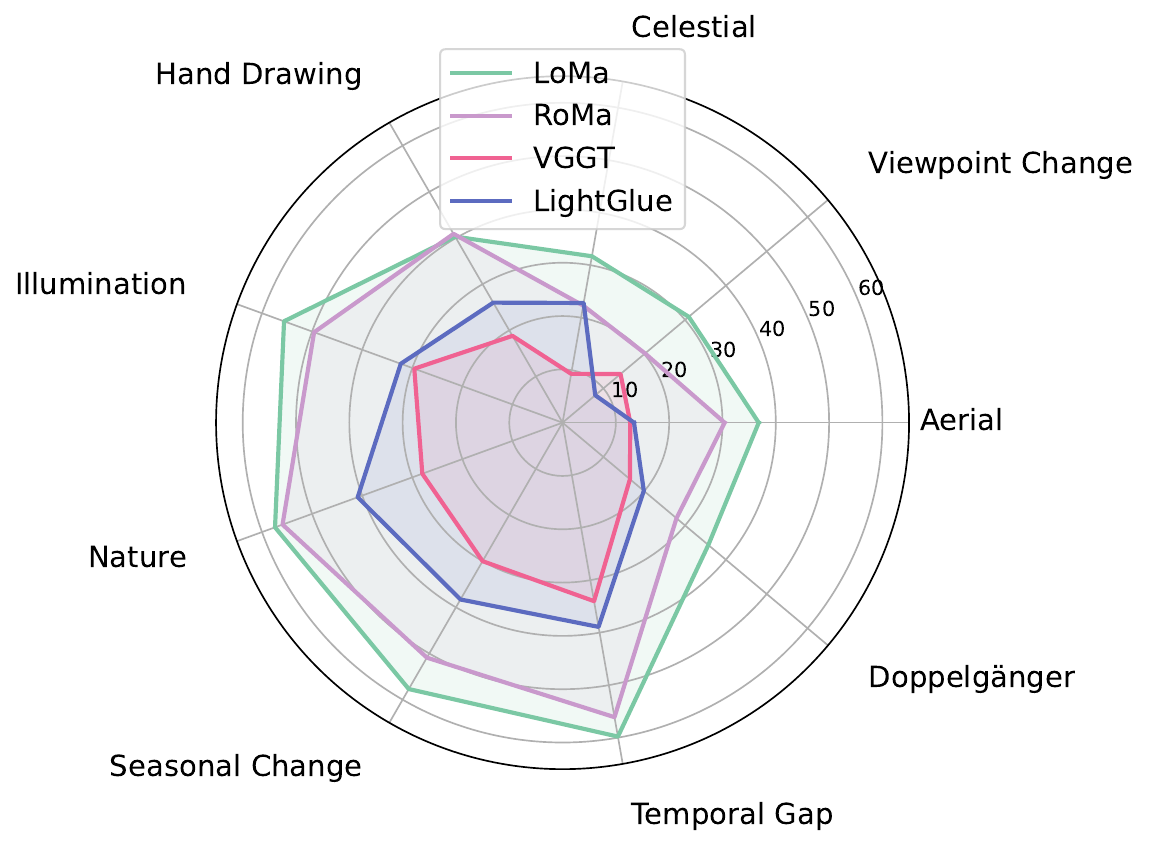}
        \caption{HardMatch mAA@10px per group.}
        \label{fig:radar}
    \end{subfigure}

    \caption{\textbf{HardMatch groups.} The dataset contains image pairs from a wide range of challenging scenarios, organized into 9 groups. (a) Example pairs illustrating each group. (b) HardMatch mAA@10px performance per group.}
    
    \label{fig:hardmatch}
\end{figure}

\section{Experiments}
\label{sec:experiments}
We compare the \ours~family of models to a wide range of sparse, dense, and feed-forward reconstruction methods on a large collection of benchmarks for extreme matching (\cref{subsec:extreme-matching}), relative pose estimation (\cref{subsec:relpose}), visual localization (\cref{subsec:visloc}), and more (\cref{subsec:additional-evals}). Finally, we give insights into ablations, throughput, and scaling (\cref{subsec:analysis}). All evaluations use $N=4096$ keypoints. See supplementary \cref{tab:varying-keypoints} for varying number of keypoints.
Additional experiments (\cref{append:additional-experiments}) and details (\cref{appendix:detail-on-evaluation}) can be found in the supplementary. We use the abbreviations SG (SuperGlue), LG (LightGlue), and SP (SuperPoint) throughout.

\subsection{Extreme Matching}\label{subsec:extreme-matching}

\subsubsection{WxBS.} We evaluate on the challenging matching benchmark WxBS~\cite{mishkin2015WXBS}. The benchmark features 37 pairs of hand-labeled correspondences that display a mix of extreme changes in viewpoint, illumination, and modality. We report the mean accuracy in \cref{tab:relpose}, and the accuracy as a function of the threshold in the supplementary (\cf \cref{fig:wxbs-curve}). \ours-G achieves SotA results on WxBS, barely beating RoMa (73.4 vs. 72.6) while handily beating other sparse matchers.

\subsubsection{HardMatch.} 
\begin{figure}[t]
    \centering
        \includegraphics[width=0.85\linewidth]{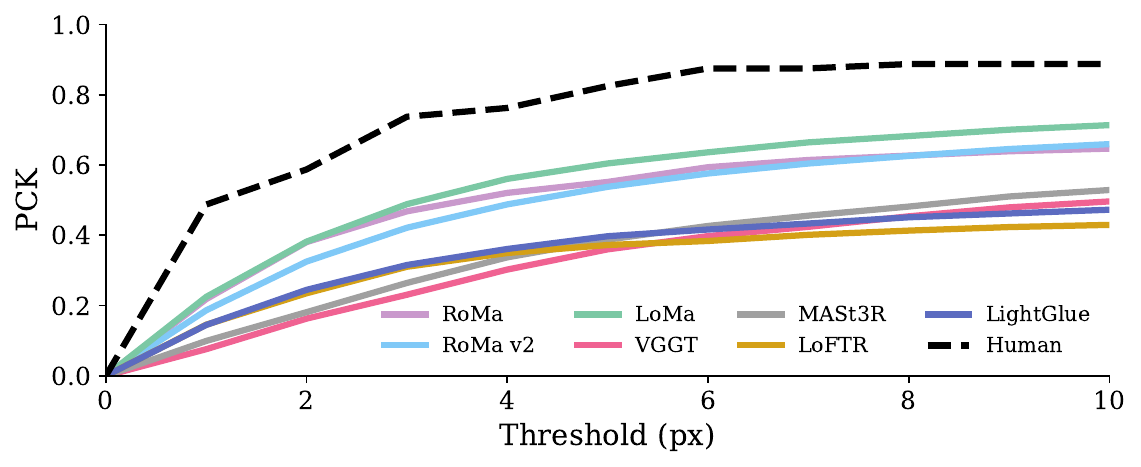}

    \caption{\textbf{HardMatch accuracy at different thresholds.} \ours~performs slightly better than the best dense matchers and significantly outperforms LightGlue.}
    \label{fig:hardmatch-curve}
\end{figure}

We report the performance on HardMatch for: (i) groups (\cref{fig:radar}), (ii) pixel thresholds (\cref{fig:hardmatch-curve}), and (iii) a wide range of matchers (\cref{tab:relpose}). The complete results are in the supplementary (\cref{tab:detailed-hardmatch}). We include a human baseline as a reference (described in \cref{subsec:data-collection}). However, the numbers are not directly comparable, as the matchers are evaluated through their estimated Fundamental matrix $F$, while the human baseline is directly evaluated on the correspondences. We find HardMatch to be challenging for SotA matchers. \ours-G achieves the best result of 54.3 mAA@10px, approximately 20 points below its performance on WxBS. Doppelgängers~\cite{cai2023doppelgangers, xiangli2025doppelgangers++}, large viewpoint changes, aerial photographs, and star constellations are particularly challenging for all matchers. We illustrate some qualitative examples in \cref{fig:qualitative-hard-groups} in the supplementary.

\subsection{Relative Pose Estimation}\label{subsec:relpose}

We compare \ours~to SotA matchers, detection+description with mutual nearest neighbor matching, and feed-forward reconstruction methods on relative pose estimation. We report the results on MegaDepth-1500~\cite{li2018megadepth, sun2021loftr} and ScanNet-1500~\cite{dai2017scannet, sarlin2020superglue} in \cref{tab:relpose}. \ours~significantly outperforms other sparse matchers on both datasets. In particular, \ours-L~achieves gains of 8.4 and 12.9 AUC@5$^\circ$ compared to other sparse matchers on MegaDepth and ScanNet, respectively.

\begin{table}[t]
\centering
\caption{\textbf{SotA matching comparison.} Relative pose estimation on MegaDepth-1500~\cite{li2018megadepth, sun2021loftr} and ScanNet-1500~\cite{dai2017scannet, sarlin2020superglue} and accuracy on WxBS~\cite{mishkin2015WXBS} and HardMatch.}
\label{tab:relpose}
\small

\begin{tabular}{l r rrr r rrr r r r r}
\toprule
Method
&& \multicolumn{3}{c}{MegaDepth}
&& \multicolumn{3}{c}{ScanNet}
&& \multicolumn{1}{c}{WxBS} && \multicolumn{1}{c}{HM} \\
\cmidrule(lr){3-5} \cmidrule(lr){7-9} \cmidrule(lr){11-11} \cmidrule(lr){13-13}
AUC$@$ $\rightarrow$ 
&& $5^{\circ}$ & $10^{\circ}$ & $20^{\circ}$
&& $5^{\circ}$ & $10^{\circ}$ & $20^{\circ}$
&mAA$@$ $\rightarrow$ & 10px && 10px \\
\midrule
\multicolumn{13}{@{}l@{}}{\small \textit{Feed-forward Reconstruction}} \\

MASt3R~\cite{leroy2024grounding}~\tiny{ECCV'24} && \bfseries 42.4 & \bfseries 61.5 & \bfseries 76.9 && 33.6 & \bfseries 56.8 & \bfseries 74.1 && 34.5 && \bfseries 33.6 \\

VGGT~\cite{wang2025vggt}~\tiny{CVPR'25} && 33.5 & 52.9 & 70.0 && \bfseries 33.9 & 55.2 & 73.4 && \bfseries 36.3 && 28.4 \\

\midrule
\multicolumn{13}{@{}l@{}}{\small \textit{Dense Matchers}} \\
LoFTR~\cite{sun2021loftr}~\tiny{CVPR'21}&& 52.8 & 69.2 & 81.2 && 22.1 & 40.8 & 57.6 && 50.7 && 33.1 \\
RoMa~\cite{edstedt2024roma}~\tiny{CVPR'24} && 62.6 & 76.7 & 86.3 && 31.8 & 53.4 & 70.9 && \bfseries 72.6 && \bfseries 48.1 \\
UFM~\cite{zhang2025ufm}~\tiny{NeurIPS'25} && 41.5 & 57.9 & 72.4 && 31.3 & 54.1 & 72.0 && 53.3 && 33.9 \\
RoMa v2~\cite{edstedt2026romav2} &&  \bfseries 62.8 &  \bfseries 77.0 & \bfseries 86.6 && \bfseries 33.6 & \bfseries 56.2 & \bfseries 73.8 && 64.8 && 46.5 \\ %
\midrule
\multicolumn{13}{@{}l@{}}{\small \textit{Detect+Describe, 4096 Keypoints}}  \\
DISK~\cite{tyszkiewicz2020disk}~\tiny{NeurIPS'20} && 35.0 & 51.4 & 64.9 && 6.4 & 13.9 & 23.2 && 21.9 && 22.0 \\  
ALIKED~\cite{Zhao2023ALIKED}~\tiny{TIM'23} && 41.9 & 58.4 & 71.7 && 6.7 & 14.6 & 25.0 && 35.1 && 26.6 \\  

DeDoDe-G~\cite{edstedt2024dedode}~\tiny{3DV'24} && 44.6 & 61.8 & 75.7 && 13.5 & 27.3 & 41.9 &&  46.4 && 30.3 \\  

\ours~Desc. (ours) && \bfseries 51.7 & \bfseries 68.3 & \bfseries 80.9 && \bfseries 18.7 & \bfseries 37.2 & \bfseries 55.6 && \bfseries 63.0 && \bfseries 39.5 \\  

\midrule
\multicolumn{13}{@{}l@{}}{\small \textit{Sparse Matchers, 4096 Keypoints}} 
\\
SP+SG~\cite{detone2018superpoint, sarlin2020superglue} && 43.7 & 61.8 & 76.5 && 16.4 & 32.5 & 49.0 && 45.6 && 36.0 \\  

SP+LG~\cite{detone2018superpoint, lindenberger2023lightglue} && 43.8 & 61.8 & 76.4 && 15.9 & 32.1 & 48.9 && 40.4 && 34.8 \\

DISK+LG~\cite{tyszkiewicz2020disk, lindenberger2023lightglue} && 47.8 & 65.3 & 79.0 && 9.3 & 19.3 & 30.8 && 39.2 && 30.3 \\
ALIKED+LG~\cite{Zhao2023ALIKED, lindenberger2023lightglue} && 48.1 & 65.7 & 79.3 && 14.5 & 28.9 & 43.5 && 43.9 && 35.7 \\

\ours-B$^{128}$~(ours) && 55.1 & 71.2 & 83.2 && 24.8 & 45.5 & 63.7 && 61.5 && 48.2 \\

\ours-B~(ours) && 55.7 & 71.8 & 83.6 && 27.5 & 49.7 & 68.2 && 68.7 && 51.1 \\

\ours-L~(ours) &&  \bfseries 56.5 & \bfseries 72.7 & \bfseries 84.3 && \bfseries 29.3 & \bfseries 51.9 & \bfseries 70.3 && 70.6 && 53.5 \\

\ours-G~(ours) && 56.1 & 72.2 & 84.0 && \bfseries 29.3 & 51.7 & 70.0 && \bfseries 73.4 && \bfseries 54.3 \\
\bottomrule
\end{tabular}
\end{table}

\subsection{Visual Localization}\label{subsec:visloc}

\subsubsection{Map-free.} The map-free relocalization benchmark~\cite{arnold2022map} tests the ability to localize the camera in metric space given a single reference image and no map. 
To obtain monocular metric depth, we use DA3~\cite{depthanything3}. Following the benchmark, we use the Virtual Correspondence Reprojection Error (VCRE<90px) and report the results for the validation set in \cref{tab:visloc}.
\ours-G~achieves a $\approx$20-point increase in precision against other sparse matchers. We provide additional comparisons in the Supplementary.

\subsubsection{InLoc.} We evaluate visual localization on InLoc~\cite{taira2018inloc} using the HLoc~\cite{sarlin2019coarse} pipeline and report the results in \cref{tab:visloc}. We find that \ours~significantly outperforms other sparse matchers. Most notably, \ours-G~achieves a more than 20-point increase over the second best matcher on the most narrow threshold for DUC2.

\subsubsection{Oxford Day-and-Night.} We evaluate visual localization under challenging lighting conditions on the Oxford Day-and-Night~\cite{wang2025seeing} dataset. In contrast to InLoc, the evaluation requires the feature matcher to construct an SfM model using the daytime database image. We use the HLoc pipeline and report the median result for night queries in \cref{tab:visloc} and individual scenes in~\cref{tab:oxford-night} in the supplementary. For the most narrow threshold, \ours-G~achieves a more than 14-point increase in accuracy compared to other sparse matchers.

\begin{table}
\centering
\caption{\textbf{Visual localization.} Comparison on Map-free~\cite{arnold2022map}, InLoc~\cite{taira2018inloc}, and Oxford Day-and-Night~\cite{wang2025seeing}. On the two latter, we report the percentage of
query images correctly localized within (0.25m, 2$^\circ$) / (0.5m, 5$^\circ$) / (1m, 10$^\circ$).}
\label{tab:visloc}
\small

\begin{tabular}{l r rr r cc r c}
\toprule
Method
&& \multicolumn{2}{c}{Map-free}
&& \multicolumn{2}{c}{InLoc}
&& \multicolumn{1}{c}{Oxford} \\
\cmidrule(lr){3-4} \cmidrule(lr){6-7} \cmidrule(lr){9-9} 

&& Prec. & AUC
&& DUC1 & DUC2
&& Night \\
\midrule
SP+SG && 46.3 & 74.1 && 46.5/65.7/78.3 & 52.7/72.5/79.4 && 44.3/54.4/58.0\\
SP+LG && 45.5 & 76.2 && 43.9/64.6/76.8 & 42.7/68.7/74.0 && 43.4/53.5/57.7\\
DISK+LG &&  43.2 & 60.7 && 43.4/60.6/74.2 & 36.6/53.4/67.2 && 14.8/17.5/20.1 \\
ALIKED+LG && 47.2 & 79.5 && 41.4/64.6/79.8 & 35.9/64.1/67.9 && 42.8/53.9/58.9 \\

\ours-B$^{128}$~(ours) && 60.8 & 87.0 && 54.5/76.8/87.9 & 64.1/82.4/84.7  && 54.8/62.2/67.1 \\

\ours-B~(ours) && 65.6 & 89.0 && 	57.1/\bfseries80.8/\bfseries91.9 & 71.0/87.0/88.5  && 54.7/62.4/66.2\\

\ours-L~(ours) && 67.6 & 89.4 && \bfseries59.1/\bfseries80.8/\bfseries91.9 & 71.0/84.0/87.8 && 56.0/64.6/69.2 \\

\ours-G~(ours) && \bfseries 68.9 & \bfseries 90.3 && 55.6/80.3/91.4 & \bfseries 73.3/87.8/89.3 && \bfseries 58.9/66.0/69.7\\

\bottomrule
\end{tabular}
\end{table}

\subsection{Additional Matching Evaluations}\label{subsec:additional-evals}

\subsubsection{RUBIK.} We further evaluate on the newly released RUBIK~\cite{loiseau2025rubik} benchmark and report the results in \cref{tab:additional-evals}. We find that \ours-G outperforms other sparse matchers, notably improving both AUC at 10$^\circ$ and 20$^\circ$ by $\approx24$ points. In \cref{fig:rubik}, we analyze how \ours~performs at different difficulty levels.

\subsubsection{Image Matching Challenge 2022.} The Image Matching Challenge (IMC) is a yearly competition held at CVPR\@. The 2022 version~\cite{image-matching-challenge-2022} consists of a hidden test-set of Google street-view images with the task to estimate the Fundamental matrix between them. \Cref{tab:additional-evals} presents the results of our submission. \ours~sets a new SoTA, handily beating other sparse matchers. \ours~also beats the competition winner from 2022 that used an ensemble of LoFTR, DKM, and SuperGlue as well as RoMa which achieved a score of $86.3$ and $88.0$, respectively. 

\begin{table}[t]
\centering
\caption{\textbf{Additional evaluations.} Comparison by AUC on  RUBIK~\cite{loiseau2025rubik} and mAA on Image Matching Challenge 2022~\cite{image-matching-challenge-2022}.}
\label{tab:additional-evals}
\small

\begin{tabular}{l r rr r r}
\toprule
Method
&& \multicolumn{2}{c}{RUBIK}
&& \multicolumn{1}{c}{IMC 2022} \\
\cmidrule(lr){3-4} 
\cmidrule(lr){6-6}   
&& @10$^\circ$ & @20$^\circ$ && @10$^\circ$ \\
\midrule
SP+SG && 46.2 & 54.1 && 72.4 \\

SP+LG && 44.8 & 52.1 && 69.2 \\

DISK+LG && 40.8 & 46.0 && 74.2 \\

ALIKED+LG && 49.0 & 55.2 && 76.9 \\

\ours-B$^{128}$~(ours) && 67.7 & 75.2 && 85.5 \\

\ours-B~(ours) && 65.7 & 72.2 && 87.4 \\

\ours-L~(ours) && 69.1 & 76.1 && 89.0   \\

\ours-G~(ours) && \bfseries 73.2 & \bfseries 79.9 && \bfseries 89.3  \\

\bottomrule
\end{tabular}
\end{table}

\subsection{Analysis}\label{subsec:analysis}

\subsubsection{Ablations.} 
\begin{figure}[t]
\centering
\begin{minipage}[t]{0.48\linewidth}
\vspace{0pt}
\centering
\small
\setlength{\tabcolsep}{4pt} %
\captionof{table}{\textbf{Ablations.} Performance on the validation set of HardMatch (HM).}
\label{tab:pipeline-ablation}

\begin{tabular}{l r r}
\toprule
Method && \multicolumn{1}{c}{HM} \\
\cmidrule(lr){3-3}
mAA$@$ $\rightarrow$ && 10px \\
\midrule
\rowcolor{gray!25} \RN{1}: ALIKED+LG (Baseline) && 36.3 \\
\RN{2}: DaD+DeDoDe+LG && 42.0 \\
\RN{3}: $\text{Matcher}\rightarrow \text{All data}$ && 48.9 \\
\RN{4}: $\text{Descriptor}\rightarrow \text{All data}$ && 51.1 \\
\RN{5}: Longer training (\ours-B) && 51.5 \\
\RN{6}: \ours-L && 52.8 \\
\rowcolor{green!25} \RN{7}: \ours-G && \bfseries 52.9 \\
\bottomrule
\end{tabular}
\end{minipage}
\hfill
\begin{minipage}[t]{0.48\linewidth}
\vspace{0pt}
\centering
\includegraphics[width=\linewidth]{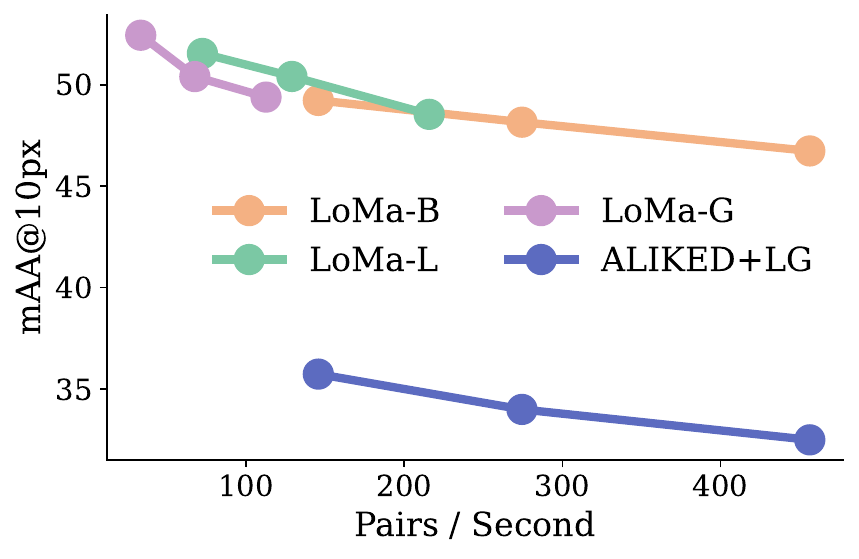}
\caption{\textbf{Pareto curve}. HardMatch performance as a function of inference speed (A100) for different stopping layers.}
\label{fig:pareto}
\end{minipage}
\end{figure}

We evaluate our design choices in \cref{tab:pipeline-ablation} by performance on the validation set of HardMatch. Changing from ALIKED to DaD+DeDoDe and retraining on MegaDepth gives a moderate boost (+5.7). Extending the training data of the matcher and subsequently also the descriptor from MegaDepth to the full dataset gives further improvements in generalization (+9.1). We then extend the training from 50K steps to 250K steps (+0.4). Scaling the matcher embedding dimension $d_{\text{emb}}$ from 256 to 1024 yields further improvements (+1.4).

\subsubsection{Throughput.}\label{subsec:inference} In SfM and visual localization, the matcher is the main bottleneck because it must process each pair of images, whereas detection and description are performed only once per image. The layer-wise loss allows the matcher to trade accuracy for speed via early stopping. We analyze this trade-off in \cref{fig:pareto} by evaluating different stopping layers ($L=\{3,5,9\}$). The \ours-B matcher has the same runtime as LG while producing significantly more accurate matches. On an A100, \ours-B can hit hundreds of pairs per second with 2048 keypoints and 16 pairs in each batch. We show the results for a single pair in each batch in \cref{fig:pareto-curve-sing-le-image} in the supplementary, which results in different model sizes being more similar in speed on modern GPUs due to poor utilization. For a single image pair, the \ours-B matcher with $L=3$ runs at almost 300 pairs per second.

\subsubsection{Scaling Local Feature Matching.}\label{subsec:scalingloma} 
\begin{figure}[t]
    \centering
    \begin{subfigure}[t]{0.45\linewidth}
        \centering
        \includegraphics[width=\linewidth]{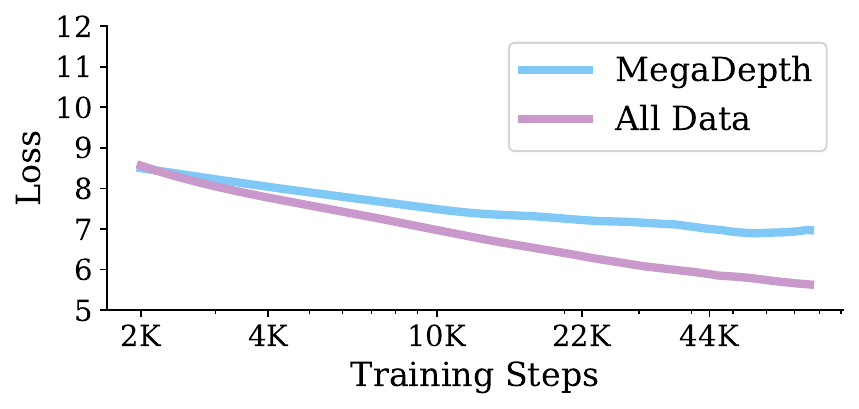}
        \caption{Data scale.}
        \label{fig:scaling-laws-data}
    \end{subfigure}
    \hfill
    \begin{subfigure}[t]{0.45\linewidth}
        \centering
        \includegraphics[width=\linewidth]{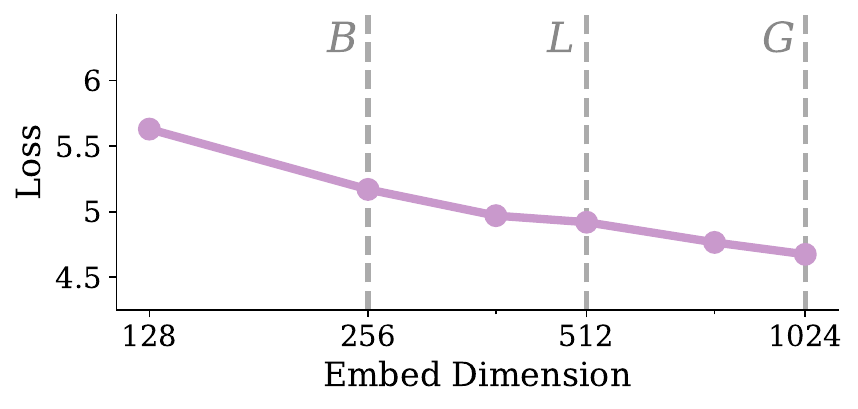}
        \caption{Model capacity.}
        \label{fig:scaling-laws-model}
    \end{subfigure}
    \caption{\textbf{Increased data scale and model capacity.} Both axes of scaling, (a) data and (b) model size, lead to significant reductions in validation loss on HardMatch.}
    \label{fig:scaling-laws}
\end{figure}

\begin{figure}[t]
    \centering
    \begin{minipage}{0.45\linewidth}
        \centering
        \includegraphics[width=\linewidth]{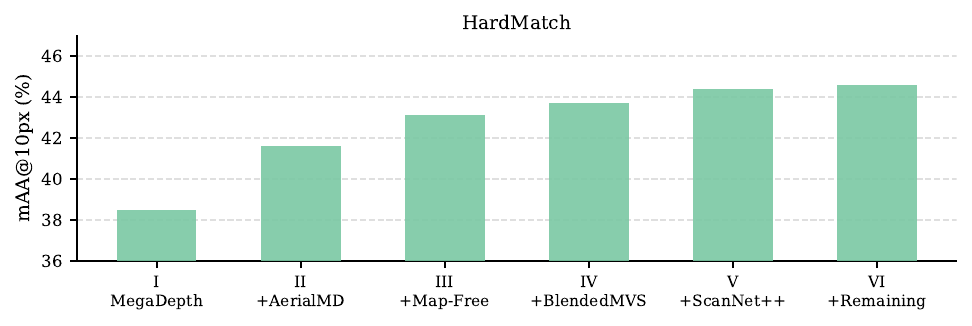}
        \caption{\textbf{Data scaling.} Effect of increased data on performance.}
        \label{fig:data-scaling}
    \end{minipage}
    \hfill
    \begin{minipage}{0.45\linewidth}
        \centering
        \includegraphics[width=\linewidth]{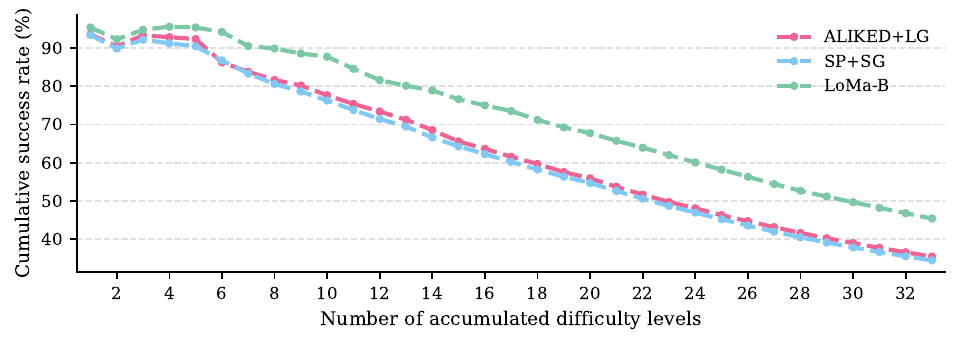}
        \caption{\textbf{RUBIK.} Cumulative success rates across difficulties.}
        \label{fig:rubik}
    \end{minipage}
\end{figure}

We find that local feature matchers benefit significantly from training on additional data (\cf \cref{fig:scaling-laws-data}) and increasing the model size (\cf \cref{fig:scaling-laws-model}). This is also illustrated through ablations in \cref{tab:pipeline-ablation}. In \cref{fig:data-scaling}, we show the performance as training data is cumulatively added (I--VI). This requires training six LoMa-B models from scratch for 100K steps each and confirms consistent gains with scale.

\section{Limitations}
Despite the strong empirical performance, 
several limitations remain.
\vspace{-5pt}
\begin{itemize}
    \item Scaling the sparse matcher works well in our experiments, but large-scale descriptor training tends to overfit, see Suppl.~\cref{fig:descriptor-scaling}. 
    \item \ours\ outperforms previous methods, but still struggles on challenging HardMatch subgroups, such as Doppelgängers and extreme viewpoint changes.
    \item HardMatch, similarly to WxBS, relies on human-annotated keypoints. The evaluation protocol based on $F$ estimation alleviates this issue, but it requires static scenes and perspective cameras.
    \item Although HardMatch is more diverse than previous benchmarks, it still contains geographic and temporal biases, see Suppl.~\cref{fig:dataset-time,fig:dataset-geography}.
    \item \ours~does not consider robustness to in-plane rotations. This can be achieved by data augmentation~\cite{nordstrom2026whohandlesorientation} or architectural design~\cite{bokman2025flopping, nordstrom2025stronger} and constitutes interesting future work.
\end{itemize}

\section{Conclusion}
We revisit the classical problem of local feature matching and show that combining large-scale data with modern practices yields substantial performance gains. To support this, we introduce (i) HardMatch, a highly challenging evaluation dataset consisting of 1000 hand-labeled image pairs, and (ii) \ours, a family of models achieving SotA performance on this new benchmark as well as on the established benchmarks IMC 2022 and WxBS, surpassing even dense matchers.

\section*{Acknowledgements}
This work was supported by the Wallenberg Artificial
Intelligence, Autonomous Systems and Software Program
(WASP), funded by the Knut and Alice Wallenberg Foundation, and by the strategic research environment ELLIIT, funded by the Swedish government. 
The computational resources were provided by the
National Academic Infrastructure for Supercomputing in
Sweden (NAISS) at C3SE, partially funded by the Swedish Research
Council through grant agreement no.~2022-06725, and by
the Berzelius resource, provided by the Knut and Alice Wallenberg Foundation at the National Supercomputer Centre.

\bibliographystyle{splncs04}
\bibliography{main}
\clearpage
\setcounter{page}{1}
\setcounter{section}{0}

\renewcommand{\thesection}{\Alph{section}}

\begin{center}
{\Large\bf \ours: Local Feature Matching Revisited}\\[0.5em]
{\large Supplementary Material}
\end{center}

\section{Additional Experiments}\label{append:additional-experiments}

\subsection{Performance using Different Detectors}

All the results in the main paper use the DaD~\cite{edstedt2025dad} detector. We investigate the performance of \ours~using different detectors in \cref{tab:changing-detector}. For fair comparison, we retrain the descriptor with an ensemble of detectors and use this for all the subsequent comparisons. To create an ensemble~\cite{wang2026understandingoptimizingattentionbasedsparse}, we uniformly sample keypoints from DeDoDe v2~\cite{edstedt2024dedodev2}, DISK~\cite{tyszkiewicz2020disk}, ALIKED~\cite{Zhao2023ALIKED}, and DaD~\cite{edstedt2025dad} during training. We then train separate matchers (one for each detector) and one ensemble matcher that is jointly trained with keypoints randomly sampled from all detectors. We find DaD to be overall the strongest detector, regardless of setting. Generally, slightly better performance is achieved by specializing the matcher on one detector.

\begin{table}[htbp]
\centering
\caption{\textbf{Detector ablation.} Performance (mAA@10px on the validation set of HardMatch) comparison between training with a single detector and randomly sampling multiple detectors (ensemble).}
\label{tab:changing-detector}
\small

\begin{tabular}{l r r r r}
\toprule
Method
&& \multicolumn{1}{c}{Ensemble Training} && \multicolumn{1}{c}{Single Detector} \\
\midrule

DeDoDe v2~\cite{edstedt2024dedodev2} && 47.0 && 47.3 \\
DISK~\cite{tyszkiewicz2020disk} && 44.3 && 45.3 \\
ALIKED~\cite{Zhao2023ALIKED} && 48.9 && 49.0 \\
DaD~\cite{edstedt2025dad} &&  \bfseries 51.0 && \bfseries 51.3 \\
\bottomrule
\end{tabular}
\end{table}

\subsection{Dependence on the Number of Keypoints}

Throughout the main paper, we evaluate with $N=4096$ keypoints. We study the performance for fewer keypoints in \cref{tab:varying-keypoints}. The performance difference between 2048 and 4096 keypoints is negligible for other sparse matchers, but \ours~benefits slightly from increasing the number of keypoints beyond 2048. Performance degrades significantly below 2048 keypoints. %

\begin{table}[htbp]
\centering
\caption{\textbf{Varying max number of keypoints.} We report the AUC@20 for different number of maximum keypoints.}
\label{tab:varying-keypoints}
\small

\begin{tabular}{l r rrrr r rrrr}
\toprule
Method
&& \multicolumn{4}{c}{MegaDepth-1500}
&& \multicolumn{4}{c}{ScanNet-1500}\\
\cmidrule(lr){3-7} \cmidrule(lr){8-11} 
Max Num. Keypoints $\rightarrow$ 
&& 512 & 1024 & 2048 & 4096 
&& 512 & 1024 & 2048 & 4096 \\
\midrule
SP+SG && 70.9 & 75.4 & 76.5 & 76.4 && 44.5 & 48.1 & 49.0 & 49.1 \\
SP+LG && 70.6 & 74.8 & 76.4 & 76.4 && 43.9 & 47.6 & 48.6 & 48.9   \\
\ours-B~(ours) &&  \bfseries 80.3 & \bfseries 82.7 & \bfseries 83.1 & \bfseries 83.6 && \bfseries 58.3 & \bfseries 63.3 & \bfseries 66.4 & \bfseries 68.2 \\
\bottomrule
\end{tabular}
\end{table}

\subsection{HPatches} 

We evaluate on HPatches~\cite{balntas2017hpatches} following the LoFTR~\cite{sun2021loftr} protocol. The dataset contains planar scenes with homographies. We report the results in \cref{tab:hpatches}. The lightweight \ours-B$^{128}$ achieves the highest score.

\begin{table}[htbp]
\centering
\caption{\textbf{HPatches.} Performance on HPatches~\cite{balntas2017hpatches}.}
\label{tab:hpatches}
\small

\begin{tabular}{l r rr r r r rrr}
\toprule
Method
&& \multicolumn{3}{c}{HPatches} \\
\cmidrule(lr){3-5} 

AUC@$\rightarrow$
&& @3px & @5px & @10px \\
\midrule
SP+SG && 64.4 & 75.6 & 86.0\\

SP+LG && 64.2 & 75.5 & 85.5 \\

DISK+LG && 60.4 & 72.4 & 83.5 \\

ALIKED+LG && 66.2 & 76.9 & 86.3 \\

\ours-B$^{128}$~(ours) && \bfseries 67.2 & \bfseries 78.1 & \bfseries 87.8 \\

\ours-B~(ours) && 66.5 & 77.5 & 87.3 \\

\ours-L~(ours) && 66.5 &  77.8 &  87.5  \\

\ours-G~(ours) && 66.4 & 77.5 & 87.3  \\

\bottomrule
\end{tabular}
\end{table}

\subsection{SatAst} 

We evaluate astronaut to satellite matching on SatAst~\cite{edstedt2026romav2}. The dataset features large in-plane rotations and scale changes, making it difficult for most matchers. We report the results in \cref{tab:additional-evals}, where we find that, while beating other sparse matchers, \ours~struggles with rotations.

\begin{table}[htbp]
\centering
\caption{\textbf{SatAst.} Astronaut to satellite matching on SatAst~\cite{edstedt2026romav2}.}
\label{tab:satast}
\small

\begin{tabular}{l r r}
\toprule
Method
&& \multicolumn{1}{c}{SatAst} \\
\cmidrule(lr){3-3} 
AUC@ $\rightarrow$
&& @10px \\
\midrule
SP+SG && 19.8 \\

SP+LG && 12.8 \\

DISK+LG && 0.0 \\

ALIKED+LG && 12.1 \\

\bfseries\ours-B~(ours) && 18.8 \\

\bfseries\ours-L~(ours) && 21.3  \\

\bfseries\ours-G~(ours) && \bfseries 22.9  \\

\bottomrule
\end{tabular}
\end{table}

\subsection{Oxford Day-and-Night}

In the main paper, we report the median performance on the night queries of Oxford Day-and-Night~\cite{wang2025seeing}. We use the HLoc~\cite{sarlin2019coarse} pipeline with NetVLAD-50~\cite{arandjelovic2016netvlad}. The benchmark contains four outdoor scenes (Bodleian Library, H.B. Allen Centre, Keble College, Observatory Quarter) and one indoor scene (Robotics Institute). In \cref{tab:oxford-night}, we report the results per scene.

\begin{table}[htbp]
\centering
\caption{\textbf{Oxford Day-and-Night.} Full results for night queries. Reporting percentage of correctly localized test images within (0.25m, 2$^\circ$) / (0.5m, 5$^\circ$) / (1m, 10$^\circ$)}
\label{tab:oxford-night}
\tiny

\begin{tabular}{l c c c c c c c c c c}
\toprule
&& \multicolumn{1}{c}{Bodleian Library}
&& \multicolumn{1}{c}{H.B. Allen Centre}
&& \multicolumn{1}{c}{Keble College}
&& \multicolumn{1}{c}{Observatory Quarter}
&& \multicolumn{1}{c}{Robotics Institute} \\
\midrule
SP+SG && 21.6/26.5/30.8 && 44.3/57.5/64.1 && 10.7/13.6/17.3 && 48.1/54.4/58.0 && 71.1/73.6/74.5  \\
SP+LG && 20.5/25.3/28.8 && 43.4/54.1/61.5 && 10.0/14.2/18.3 && 47.9/53.5/57.7 && 70.1/71.8/73.0 \\
DISK+LG && 14.8/17.5/20.1 && 9.6/11.4/14.7 && 0.5/ 0.8/1.1 && 16.8/20.4/22.9 && 53.2/57.5/60.5 \\
ALIKED+LG && 22.2/27.3/31.0 && 42.8/55.5/62.6 && 10.3/13.5/18.3 && 45.1/53.9/58.9 && 57.2/61.4/63.6 \\

\ours-B$_{128}$(ours) && \bfseries 26.1/31.8/36.1 && 60.1/71.1/76.6 && 14.2/17.8/22.0 && 54.8/62.2/67.1 && \bfseries 73.3/76.1/77.0 \\

\ours-B~(ours) && 24.1/28.9/32.5 && 57.5/69.7/73.7 && 15.3/21.0/25.2 && 54.7/62.4/66.2 && 70.9/74.1/74.7 \\

\ours-L~(ours) && 25.1/30.4/34.3 && \textbf{62.1}/71.9/76.0 && 15.4/21.1/26.0 &&  56.0/64.6/69.2  &&  71.5/74.9/76.0 \\

\ours-G~(ours) && 25.1/30.4/35.0 && 61.9/\textbf{73.7}/\textbf{78.2} && \bfseries 15.8/21.4/27.5 && \bfseries 58.9/66.0/69.7 && 72.2/75.2/76.5 \\

\bottomrule
\end{tabular}
\end{table}

\subsection{WxBS}

\begin{figure}[htbp]
    \centering
        \includegraphics[width=\linewidth]{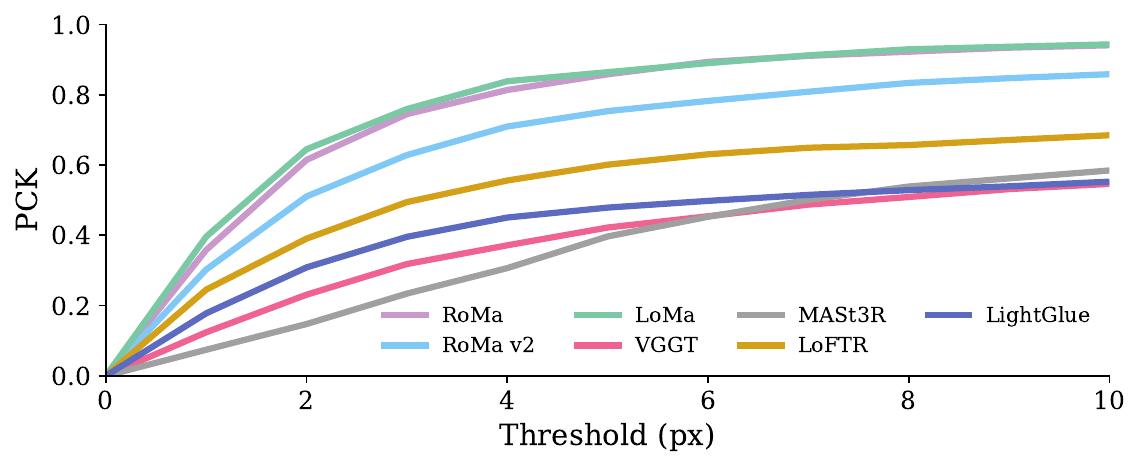}

    \caption{\textbf{WxBS accuracy at different thresholds.}}
    \label{fig:wxbs-curve}
\end{figure}

To better understand the relative performance of the matchers, we report the accuracy (PCK) at different pixel thresholds in \cref{fig:wxbs-curve}. 

\subsection{Scaling the Descriptor}

\begin{figure}[htbp]
    \centering
    \begin{subfigure}[t]{0.49\linewidth}
        \includegraphics[width=\linewidth]{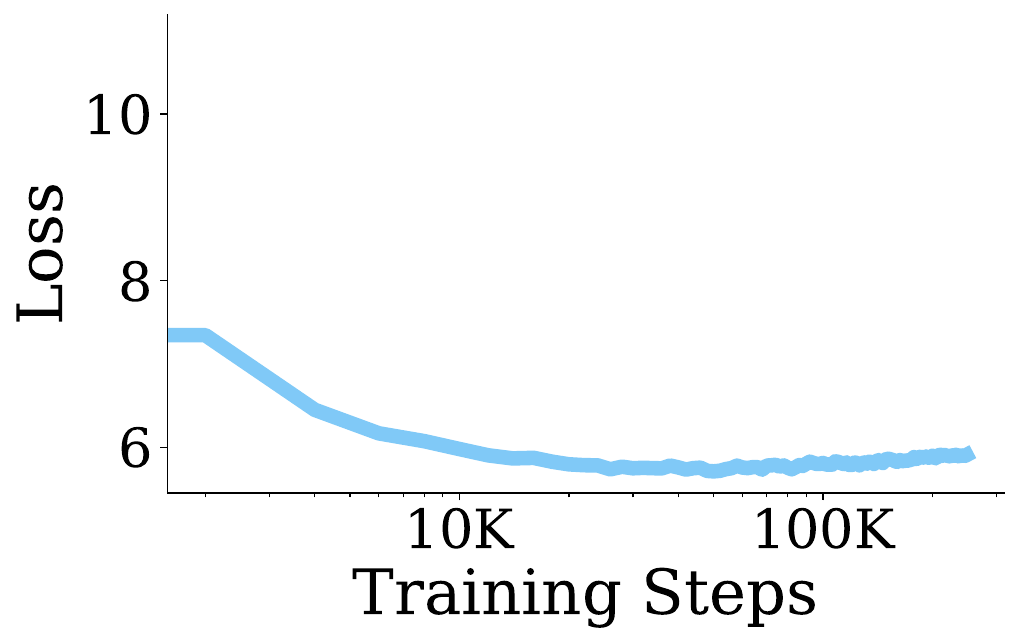}
        \caption{HardMatch validation loss during descriptor training.}
        \label{fig:descriptor-scaling}
\end{subfigure}%
    \hfill 
    \begin{subfigure}[t]{0.49\linewidth}
\includegraphics[width=\linewidth]{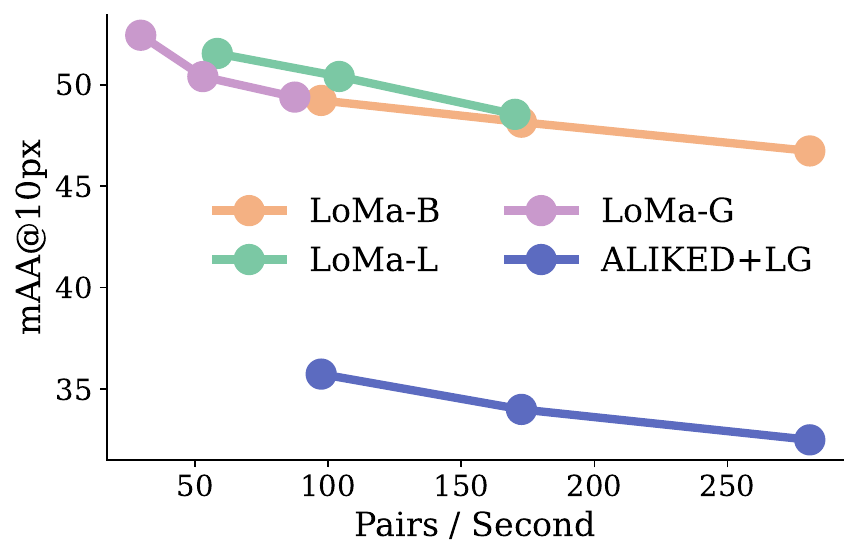}
        \caption{Pareto curve for a single image pair per GPU.}
        \label{fig:pareto-curve-sing-le-image}
\end{subfigure}
\caption{Descriptor scaling (a) and pareto curve for batch size of 1 (b).}
\end{figure}

As shown in \cref{fig:descriptor-scaling}, the HardMatch validation loss of the descriptor saturates at around 50K steps and then slowly increases. Thus, we limit  the descriptor training to only 50K steps (compared to 250K for the matcher).

\subsection{Inference Speed for a Single Image Pair}\label{subsec:single-image-inference}

In the main paper, we evaluate the speed using a batch size of 16. In many applications, inference will run with a single image pair at a time (batch size of 1). In \cref{fig:pareto-curve-sing-le-image}, we show the speed for our different model sizes for different stopping layers $L=\{3, 5, 9\}$.

\subsection{Map-free Visual Relocalization Benchmark}\label{appendix:map-free}

In \cref{tab:map-free}, we compare to additional matchers on the validation set (using DA3 depths for consistency). We also evaluated LoMa-G on the official benchmark, achieving an AUC and precision (VCRE$<$90px) of 92.5 and 75.7, respectively.

\begin{table}[htbp]
\centering
\caption{\textbf{Additional Map-free Baselines.} TBD}
\label{tab:map-free}

\begin{tabular}{lcccc}
\toprule
Metric & MASt3R & LoFTR & RoMa & LoMa-G \\
\midrule
Prec. $\uparrow$ & 73.4 & 39.7 & 59.7 & 68.9 \\
AUC $\uparrow$   & 91.6 & 73.5 & 84.4 & 90.3 \\
\bottomrule
\end{tabular}
\end{table}
\begin{center}

\end{center}

\section{Details on Evaluation}\label{appendix:detail-on-evaluation}

For all \ours~evaluations, we use an internal resolution of $784\times 784$, $N=4096$ DaD keypoints, and $L=9$ layers.

\subsection{Relative Pose Estimation}
The evaluation protocol follows from LoFTR~\cite{sun2021loftr} and is also used in \eg RoMa~\cite{edstedt2024roma} and RoMa v2~\cite{edstedt2026romav2}. We use a RANSAC pixel threshold of $\tau=0.5$. We use the standard AUC metric. The AUC metric evaluates the error of the estimated Essential matrix relative to the ground truth. For each image pair, the error is defined as the maximum of the rotational and translational errors. Since metric scale is unavailable, the translational error is measured using the cosine of the angular difference. The recall at a threshold $\tau$ is defined as the fraction of pairs whose error is below $\tau$. The metric $\mathrm{AUC}@\tau^\circ$ is computed as the normalized integral of the recall curve with respect to the threshold from $0$ to $\tau$, divided by $\tau$. In practice, this integral is approximated using the trapezoidal rule over the set of errors produced by the method on the dataset.

\subsection{Image Matching Challenge 2022}

We run the official Kaggle competition and use 200K iterations of MAGSAC~\cite{barath2019magsac, barath2020magsac++} with an inlier threshold of $\tau=0.2$. We report the mean average accuracy (mAA). The metric evaluates the estimated Fundamental matrix against the hidden ground truth using rotational error (in degrees) and translational error (in meters). A pose is considered correct if both errors fall below specified thresholds. This is evaluated over ten uniformly spaced threshold pairs. The mAA is the average accuracy across all thresholds and images, balanced across scenes.

\section{HardMatch}\label{appendix:hardmatch}

\subsection{Further Details on Evaluation}\label{appendix:hardmatch-evaluation}

Following WxBS~\cite{mishkin2015WXBS}, our main method for evaluating the hand-labeled correspondences in HardMatch is through the estimation of a Fundamental matrix. Specifically, each method finds matches between the two images and robustly estimates the fundamental matrix using the OpenCV implementation of MAGSAC~\cite{barath2019magsac, barath2020magsac++} with an inlier threshold of $\tau=0.25$ pixels. We compute the percent of keypoints (PCK) in the ground truth correspondences consistent with the estimated Fundamental matrix for epipolar pixel thresholds going from 0 to 20. We compute the pixel errors at a resolution of $640 \times 640$ and do not evaluate on the approximately 20 pairs we label dynamic.

\subsection{Correspondence Evaluation}\label{appendix:correspondence-evaluation}

An alternative methodology involves directly matching the ground truth keypoints. 
This has the benefit of working even for image pairs where a Fundamental matrix is not well defined, \eg dynamic scenes and non-perspective cameras. 
For dense matchers,we sample the warp at the ground truth keypoint locations in $I^{\mathcal{A}}$ and find the pixel error to the ground truth correspondences in $I^{\mathcal{B}}$. 
For sparse matchers, the most straightforward way is to append the ground truth keypoints to the detected keypoints in $I^{\mathcal{A}}$ and record the pixel error between the estimated and true matches in $I^{\mathcal{B}}$. As illustrated in \cref{tab:correspondence-eval}, \ours~performs the best also on this evaluation.

\begin{table}[htbp]
\centering
\caption{\textbf{Correspondence Evaluation.} HardMatch evaluation where, for sparse matchers, ground truth correspondences are appended to detected keypoints.}
\label{tab:correspondence-eval}
\small

\begin{tabular}{l r rrrr}
\toprule
Method
&& \multicolumn{4}{c}{HardMatch} \\
\cmidrule(lr){3-6} 
PCK$@$ $\rightarrow$
&& 5px & 10px & 15px & 20px \\

\midrule
\multicolumn{6}{@{}l@{}}{\small \textit{Dense Matchers}} \\
RoMa && 52.9 & 60.9 & 64.1 & 66.0\\
UFM && 35.2 & 47.1 & 53.4 & 56.7 \\
RoMa v2 && \bfseries 53.2 & \bfseries 64.5 & \bfseries 70.4 & \bfseries 73.3 \\

\midrule
\multicolumn{6}{@{}l@{}}{\small \textit{Sparse Matchers, 2048 keypoints}} \\
SP+SG && 37.6 & 39.3 & 40.2 & 40.6 \\
SP+LG && 39.3 & 45.1 & 47.8 & 49.5 \\

\ours-B~(ours) && 64.3 & 71.4 & 74.5 & 76.0\\
\ours-L~(ours) && 65.8 & 72.7 & 76.0 & 77.4 \\
\ours-G~(ours) && \bfseries 68.0 & \bfseries 74.0 & \bfseries 76.8 & \bfseries 78.2 \\

\bottomrule
\end{tabular}
\end{table}

\subsection{Dataset statistics}

\begin{figure}[htbp]
    \centering
    \begin{subfigure}[t]{0.49\linewidth}
        \includegraphics[width=\linewidth]{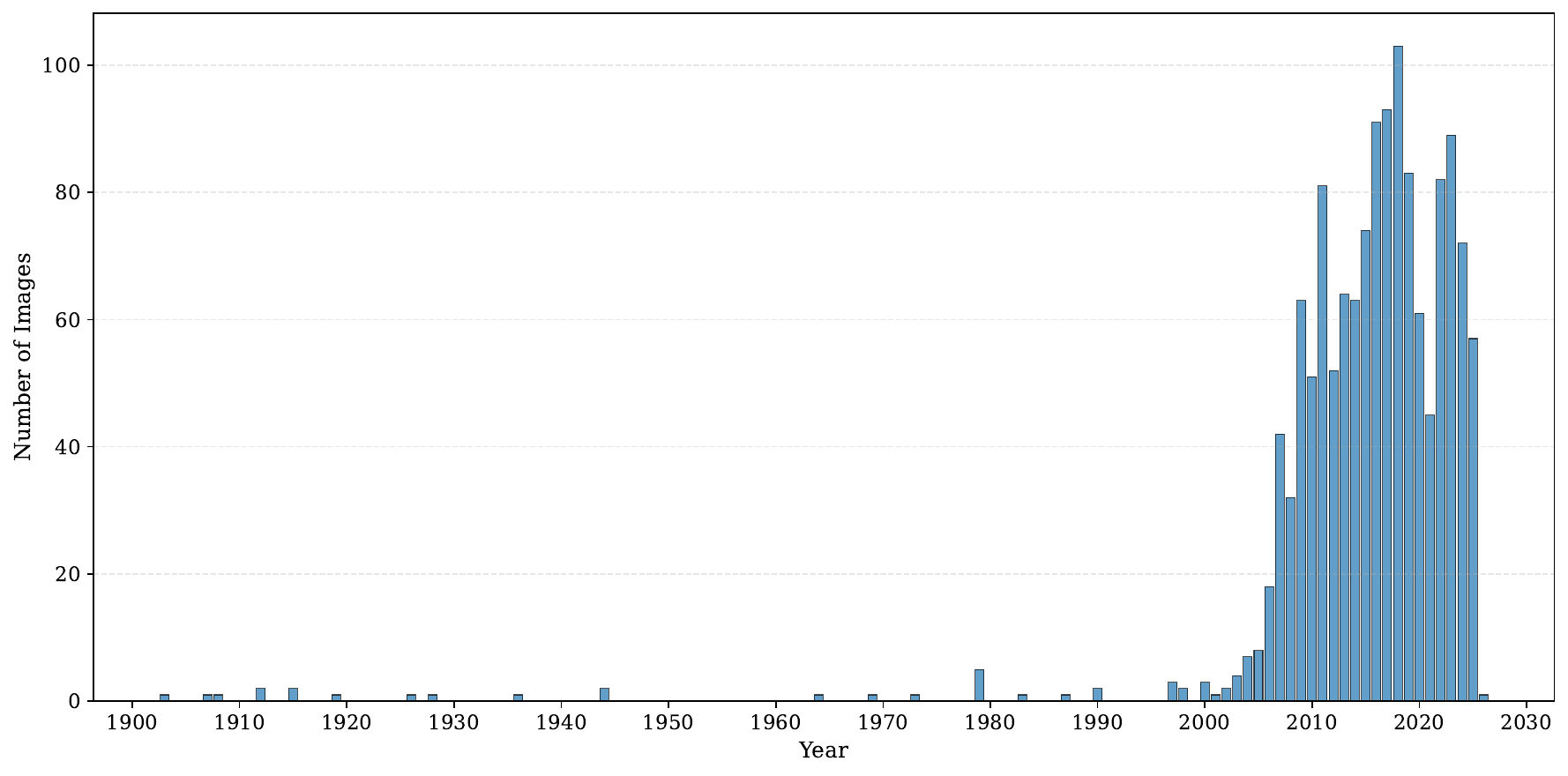}
        \caption{Time distribution}
        \label{fig:dataset-time}
\end{subfigure}%
    \hfill 
    \begin{subfigure}[t]{0.49\linewidth}
\includegraphics[width=\linewidth]{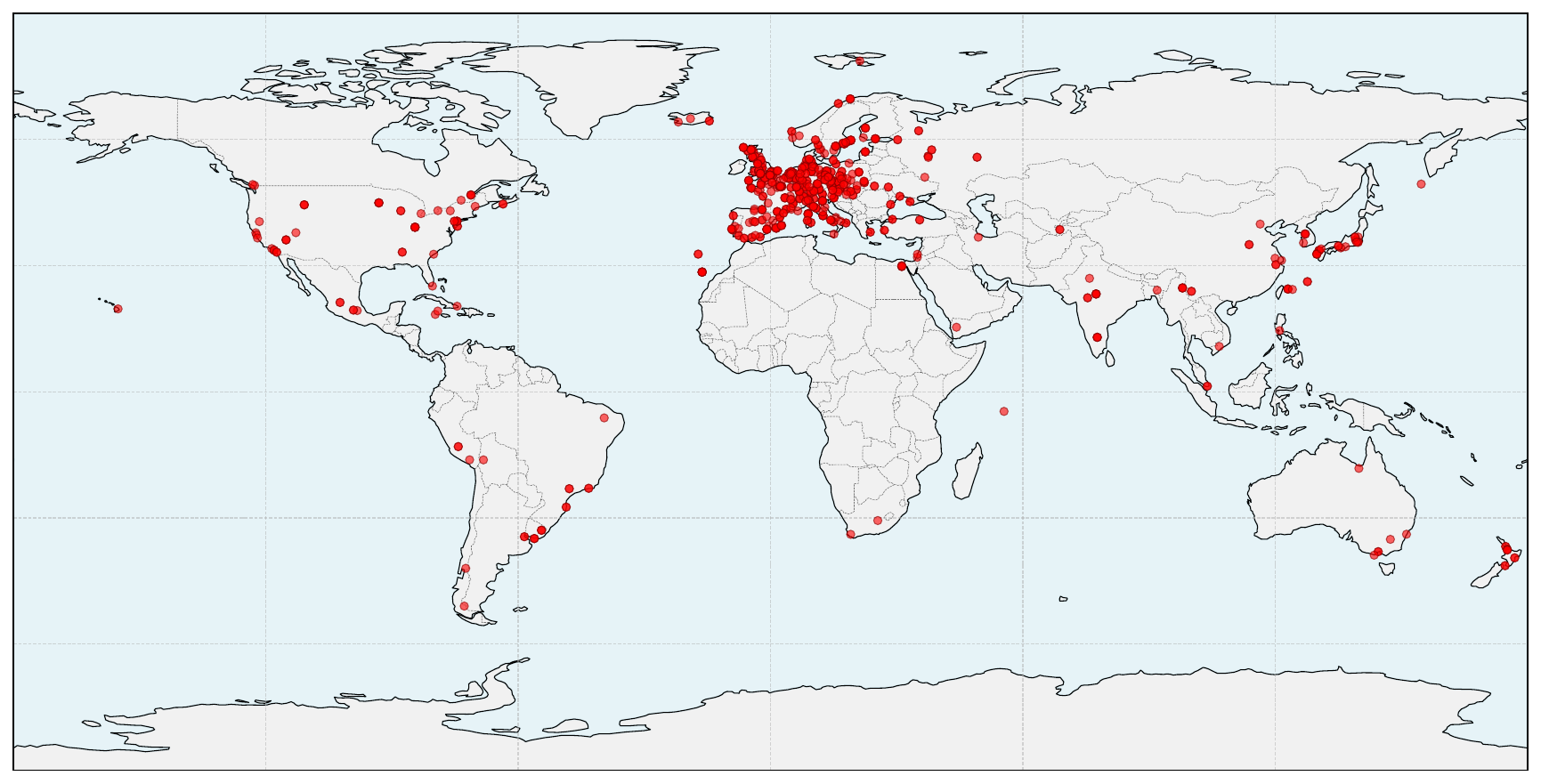}
        \caption{Geographic distribution}
        \label{fig:dataset-geography}
\end{subfigure}
\caption{\textbf{HardMatch statistics.} The dataset consists of images taken from all over the world and from over a century apart. The highest concentration is geographically in Europe and temporally in the 21st century.}
    \label{fig:hardmatch-statistics}
\end{figure}

Images in the HardMatch dataset date between the early 20th century until now (\cf \cref{fig:dataset-time}) and has a global geographic footprint (\cf \cref{fig:dataset-geography}). Most of the images are taken at the start of the 21st century in Europe. 

\subsection{Qualitative Pairs with Matches}

\begin{figure}[htbp]
    \centering
        \includegraphics[width=\linewidth]{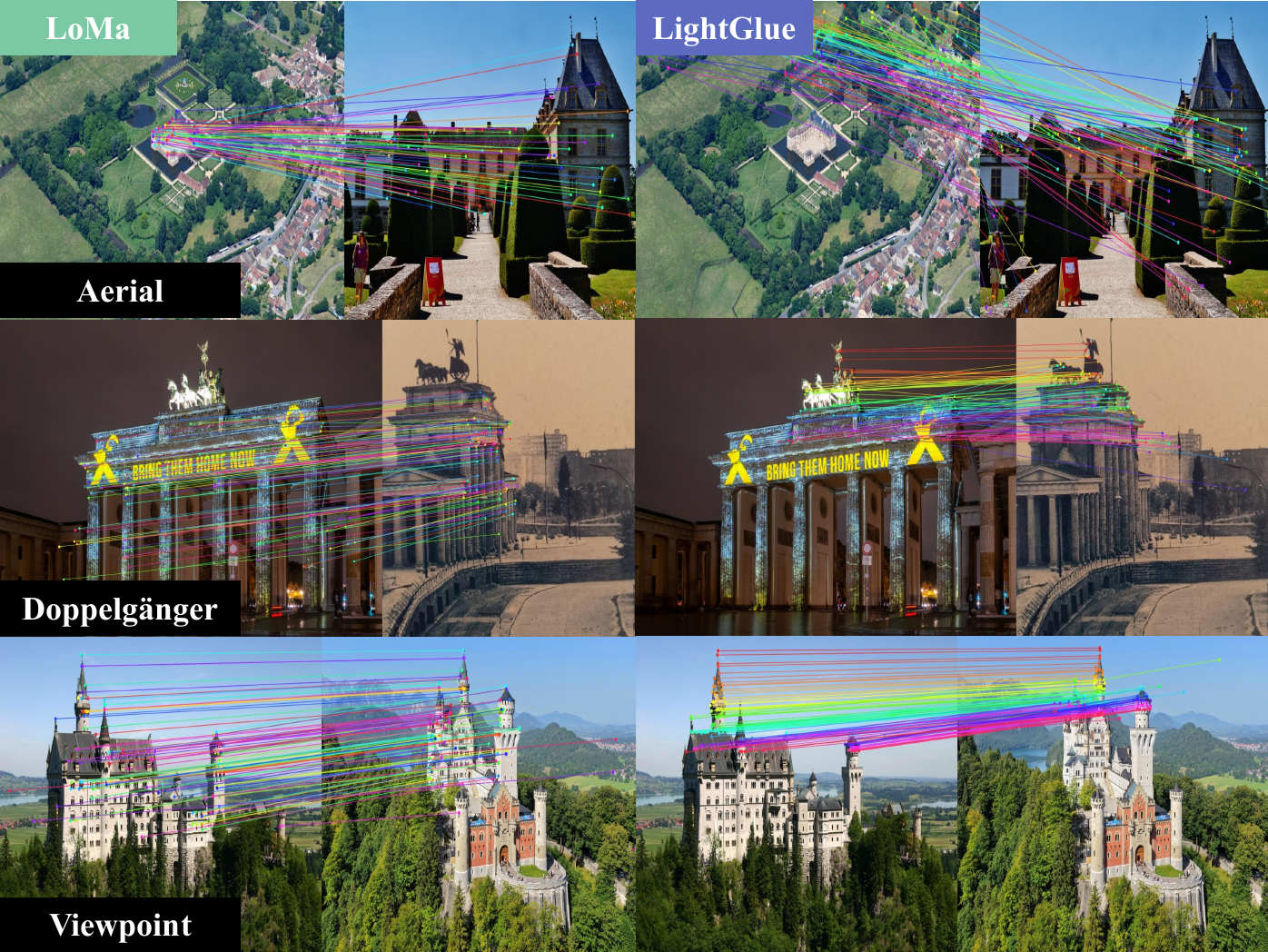}

    \caption{\textbf{Hard groups of HardMatch.} For hard Doppelgängers in HardMatch, all the matchers fail.}
    \label{fig:qualitative-hard-groups}
\end{figure}

\begin{figure}[htbp]
    \centering
    \includegraphics[width=0.33\linewidth]{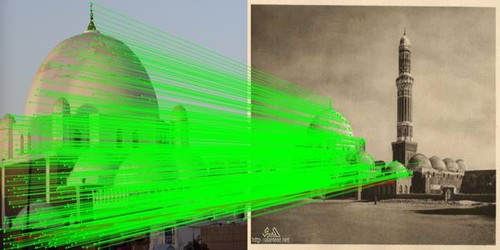}\hfill%
    \includegraphics[width=0.33\linewidth]{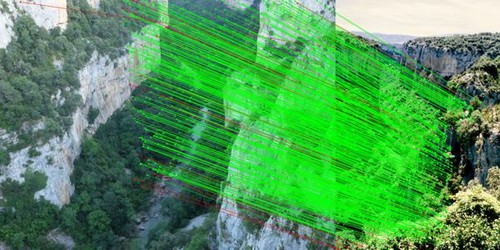}\hfill%
    \includegraphics[width=0.33\linewidth]{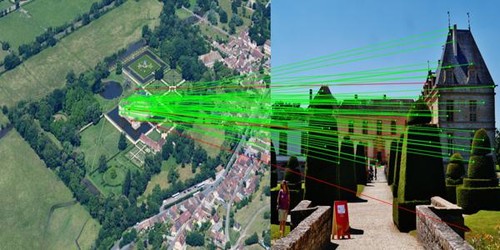}\hfill

    \includegraphics[width=0.33\linewidth]{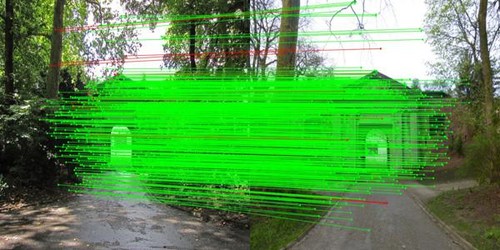}\hfill%
    \includegraphics[width=0.33\linewidth]{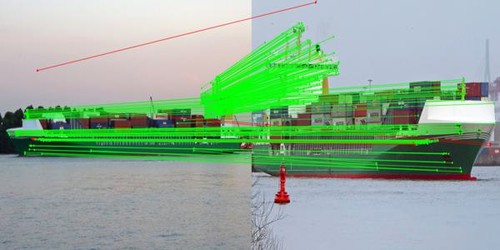}\hfill%
    \includegraphics[width=0.33\linewidth]{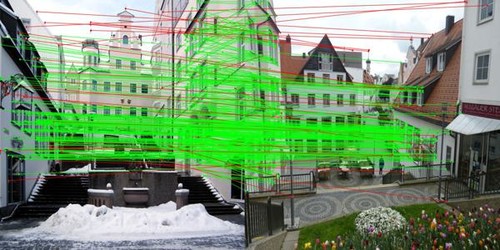}\hfill

    \includegraphics[width=0.33\linewidth]{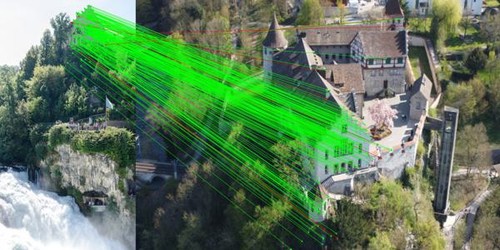}\hfill%
    \includegraphics[width=0.33\linewidth]{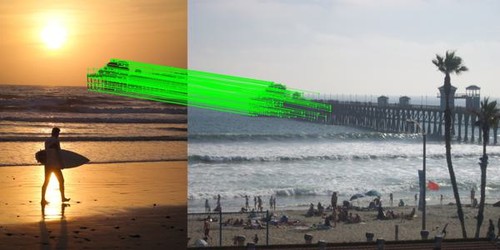}\hfill%
    \includegraphics[width=0.33\linewidth]{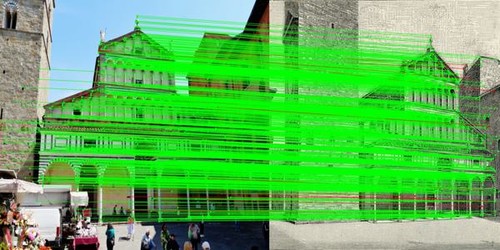}\hfill

    \includegraphics[width=0.33\linewidth]{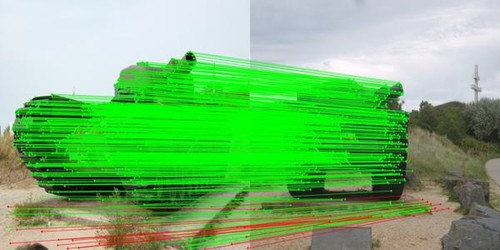}\hfill%
    \includegraphics[width=0.33\linewidth]{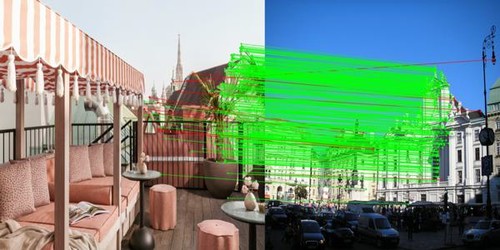}\hfill%
    \includegraphics[width=0.33\linewidth]{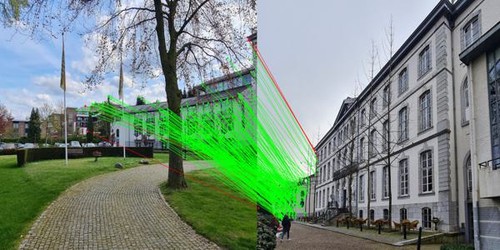}\hfill

    \includegraphics[width=0.33\linewidth]{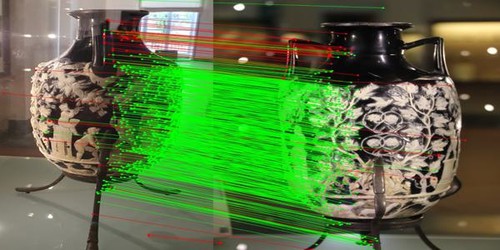}\hfill%
    \includegraphics[width=0.33\linewidth]{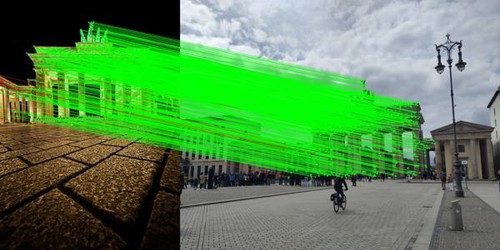}\hfill%
    \includegraphics[width=0.33\linewidth]{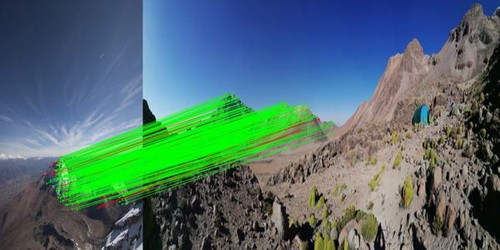}\hfill

    \includegraphics[width=0.33\linewidth]{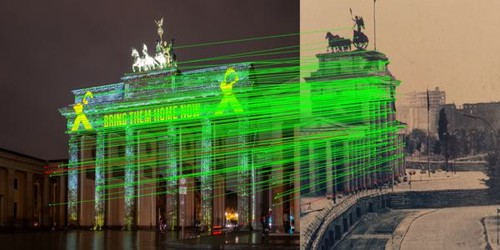}\hfill%
    \includegraphics[width=0.33\linewidth]{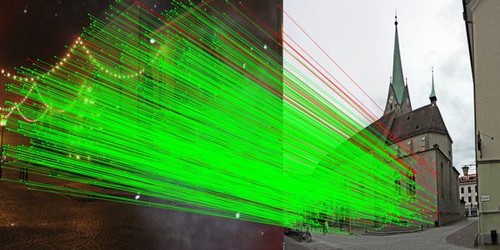}\hfill%
    \includegraphics[width=0.33\linewidth]{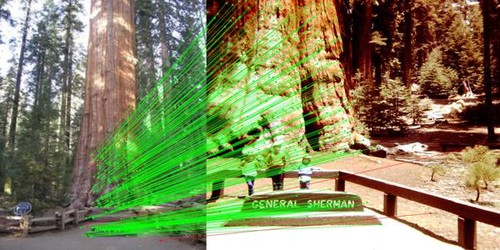}\hfill

    \includegraphics[width=0.33\linewidth]{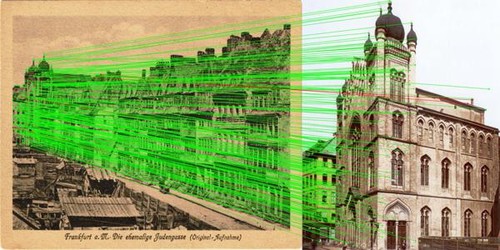}\hfill%
    \includegraphics[width=0.33\linewidth]{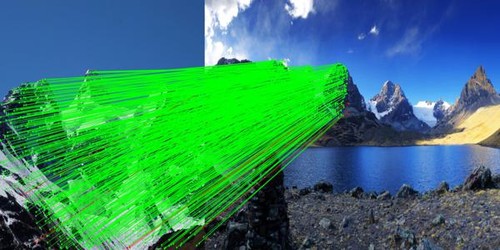}\hfill%
    \includegraphics[width=0.33\linewidth]{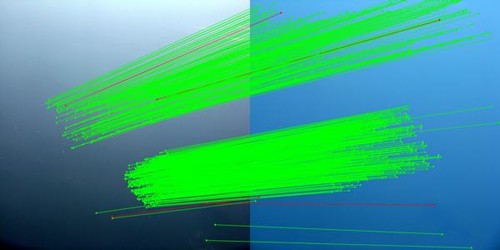}\hfill

    \includegraphics[width=0.33\linewidth]{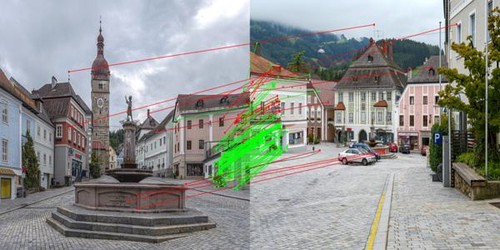}\hfill%
    \includegraphics[width=0.33\linewidth]{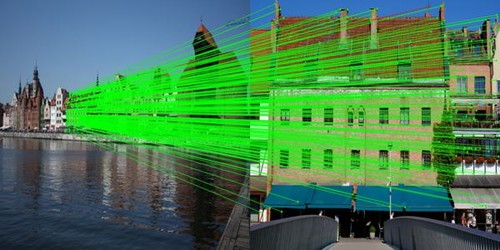}\hfill%
    \includegraphics[width=0.33\linewidth]{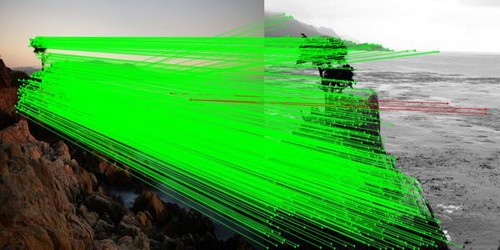}\hfill
    
    \caption{\textbf{\ours-G~matches from HardMatch.} Inliers at 5px threshold for MAGSAC~\cite{barath2019magsac, barath2020magsac++} colored green while outliers are colored red.}
    \label{fig:hardmatch-pairs-qual}
    \vspace{-1.5em}
\end{figure}

In \cref{fig:qualitative-hard-groups} we illustrate some representative examples for challenging groups in HardMatch. In \cref{fig:hardmatch-pairs-qual} we display a random collection of pairs and the matches detected by \ours-G (inliers during Fundamental matrix estimation with $\tau=5$ are colored green).

\subsection{Results by Category and Group}

\begin{figure}[htbp]
    \centering
    \includegraphics[width=0.8\linewidth]{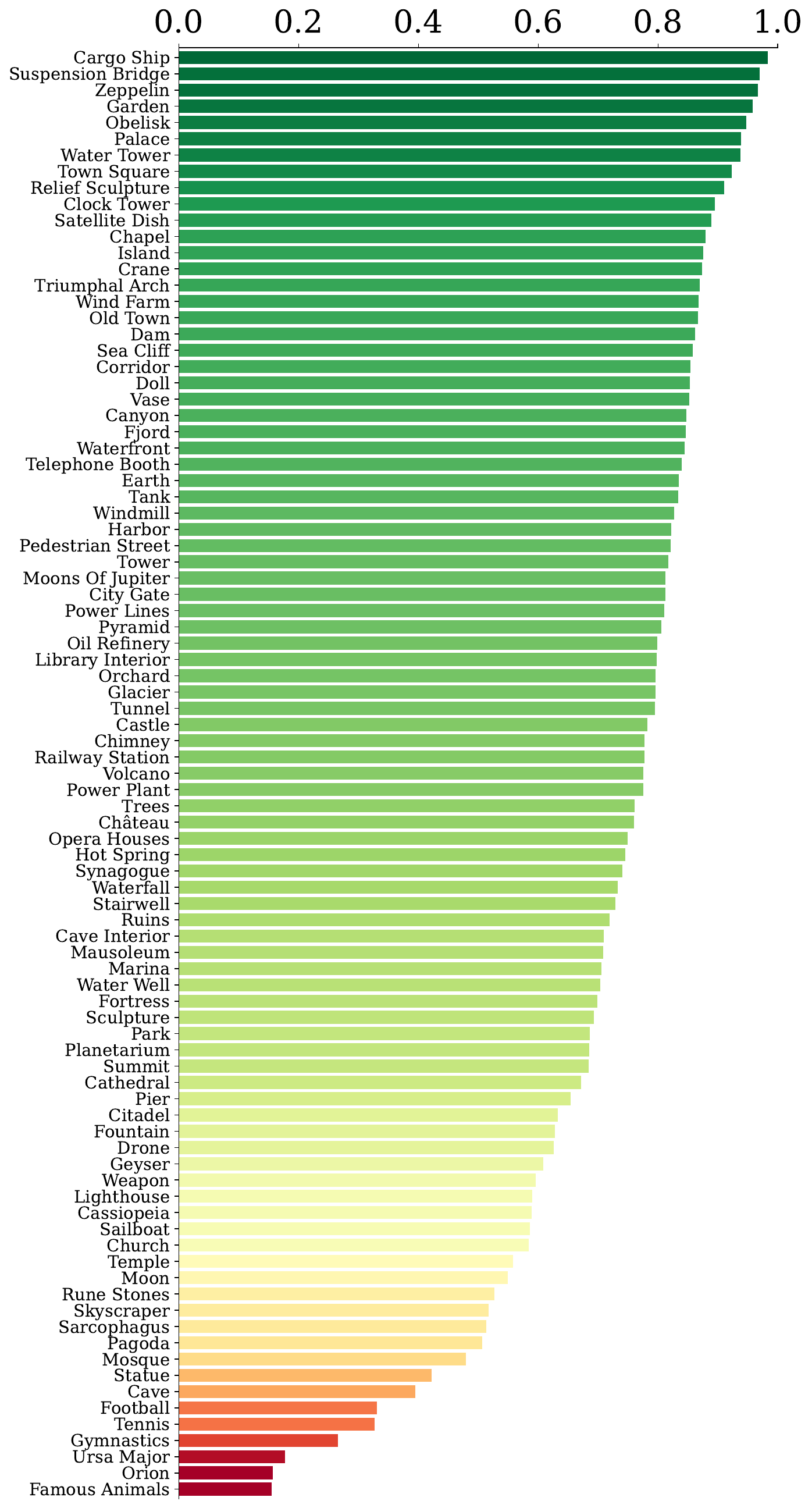}
        \caption{\textbf{HardMatch categories from easy to hard.} We plot the PCK@10px of \ours~for different categories in the test set.}
        \label{fig:categories-distribution}
\end{figure}

The HardMatch dataset is sourced from 100 Wikimedia Commons categories. We list the categories of the test dataset in \cref{fig:categories-distribution}. We also report the detailed performance breakdown of different groups in \cref{tab:detailed-hardmatch}. 

\setcounter{table}{13}
\begin{table}[htbp]
\centering
\caption{\textbf{Detailed HardMatch Performance.} Performance (mAA@10px) on different groupings of HardMatch.}
\label{tab:detailed-hardmatch}
\tiny
\begin{tabular}{l r r rrrrrrrrrrr }
\toprule
Method 
&& Aerial & Celestial & Doppelgänger & Drawing & Illumination & Nature & Seasonal & Temporal & Viewpoint \\
\midrule
\multicolumn{11}{@{}l@{}}{\tiny \textit{Feedforward Reconstruction}} \\

MASt3R~\cite{leroy2024grounding} && \bfseries 13.0 & \bfseries 20.2 & \bfseries 25.4 & \bfseries 29.1 & \bfseries 32.3 & \bfseries 39.5 & \bfseries 29.7 & \bfseries 32.3 & \bfseries 18.2 \\
VGGT~\cite{wang2025vggt} && 12.9 & 8.7 & 14.8 & 17.9 & 32.2 & 27.7 & 30.6 & 32.0 & 15.6 \\
\midrule
\multicolumn{11}{@{}l@{}}{\tiny \textit{Dense Matchers}} \\

LoFTR~\cite{sun2021loftr} && 12.9 & 25.9 & 19.3 & 21.3 & 33.5 & 39.5 & 34.3 & 38.6 & 10.5 \\
RoMa~\cite{edstedt2024roma} && 27.2 & 26.1 & 28.7 & \bfseries 41.2 & \bfseries 50.0 & 54.2 & 51.3 & \bfseries 55.0 & 20.8  \\
UFM~\cite{zhang2025ufm} && 14.4 & \bfseries 30.7 & 22.4 & 30.4 & 32.0 & 33.0 & 40.3 & 41.7 & 15.0 \\
RoMa v2~\cite{edstedt2026romav2} && \bfseries 28.7 & 28.3 & \bfseries 34.1 & 34.3 & 49.1 & \bfseries 54.4 & \bfseries 46.5 & 50.5 & \bfseries 28.6 \\

\midrule
\multicolumn{11}{@{}l@{}}{\tiny \textit{Sparse Matchers, 4096 Keypoints}} \\

SP+SG~\cite{detone2018superpoint, sarlin2020superglue} && 16.7 & 28.1 & 23.5 & 27.2 & 34.5 & 41.5 & 42.2 & 40.8 & 8.6 \\  

SP+LG~\cite{detone2018superpoint, lindenberger2023lightglue} && 12.9 & 25.1 & 22.1 & 26.8 & 32.8 & 40.0 & 37.2 & 39.2 & 8.3   \\

DISK+LG~\cite{tyszkiewicz2020disk, lindenberger2023lightglue} && 12.0 & 6.3 & 16.8 & 20.0 & 25.3 & 30.7 & 29.2 & 36.9 & 9.6  \\
ALIKED+LG~\cite{Zhao2023ALIKED, lindenberger2023lightglue} && 14.0 & 16.4 & 21.5 & 26.0 & 34.2 & 40.9 & 39.1 & 41.6 & 10.5  \\

\ours-B~(ours) && 31.6 & 28.1 & 30.3 & 41.0 & 51.7 & 54.4 & 54.6 & 55.6 & 26.0 \\

\ours-L~(ours) && 35.6 & \bfseries 35.8 & 35.9 & \bfseries 43.0 & 54.5 & 55.2 & 56.4 & 57.6 & 30.3 \\

\ours-G~(ours) && \bfseries 36.9 & 35.7 & \bfseries 36.3 & 40.4 & \bfseries 55.0 & \bfseries 56.0 & \bfseries 58.8 & \bfseries 59.4 & \bfseries 30.9  \\

\bottomrule
\end{tabular}
\end{table}

\section{Progressive Match Refinement}

\begin{figure}[htbp]
    \centering
    
    \begin{subfigure}{0.49\linewidth}
        \centering
        \includegraphics[width=\linewidth]{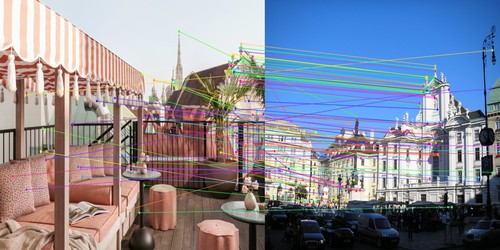}
        \caption{$L=0$}
    \end{subfigure}
    \begin{subfigure}{0.49\linewidth}
        \centering
        \includegraphics[width=\linewidth]{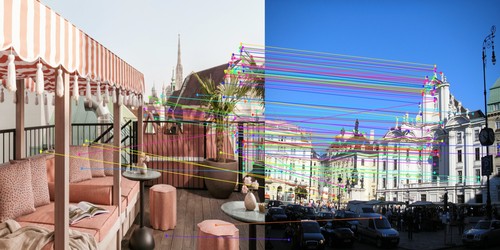}
        \caption{$L=1$}
    \end{subfigure}

    \vspace{0.5em}

    \begin{subfigure}{0.49\linewidth}
        \centering
        \includegraphics[width=\linewidth]{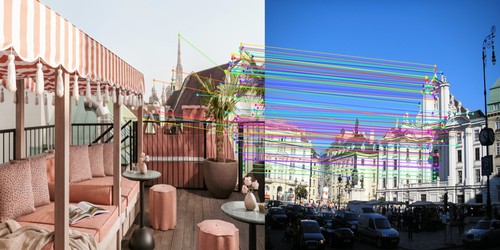}
        \caption{$L=3$}
    \end{subfigure}
    \begin{subfigure}{0.49\linewidth}
        \centering
        \includegraphics[width=\linewidth]{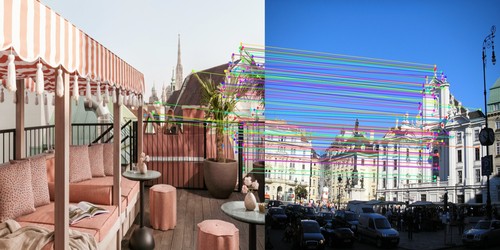}
        \caption{$L=5$}
    \end{subfigure}

    \vspace{0.5em}

    \begin{subfigure}{0.49\linewidth}
        \centering
        \includegraphics[width=\linewidth]{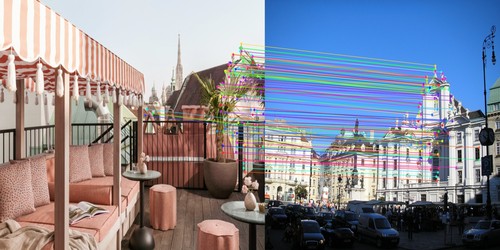}
        \caption{$L=7$}
    \end{subfigure}
    \begin{subfigure}{0.49\linewidth}
        \centering
        \includegraphics[width=\linewidth]{figures/progressive-matching/LoMa_G.jpg}
        \caption{L=9}
    \end{subfigure}

    \caption{\textbf{Refining matches through depth.} The descriptor fails to match the pair ($L=0$) but as the features are passed through the layers of the matcher, the pair gradually becomes matchable.}
    \vspace{-1.5em}
    \label{fig:refining-matches}
\end{figure}

We qualitatively examine the detected matches at different stopping layers in \cref{fig:refining-matches}.

\section{Visualizing a Training Batch}

\begin{figure}[htbp]
    \centering
        \includegraphics[width=\linewidth]{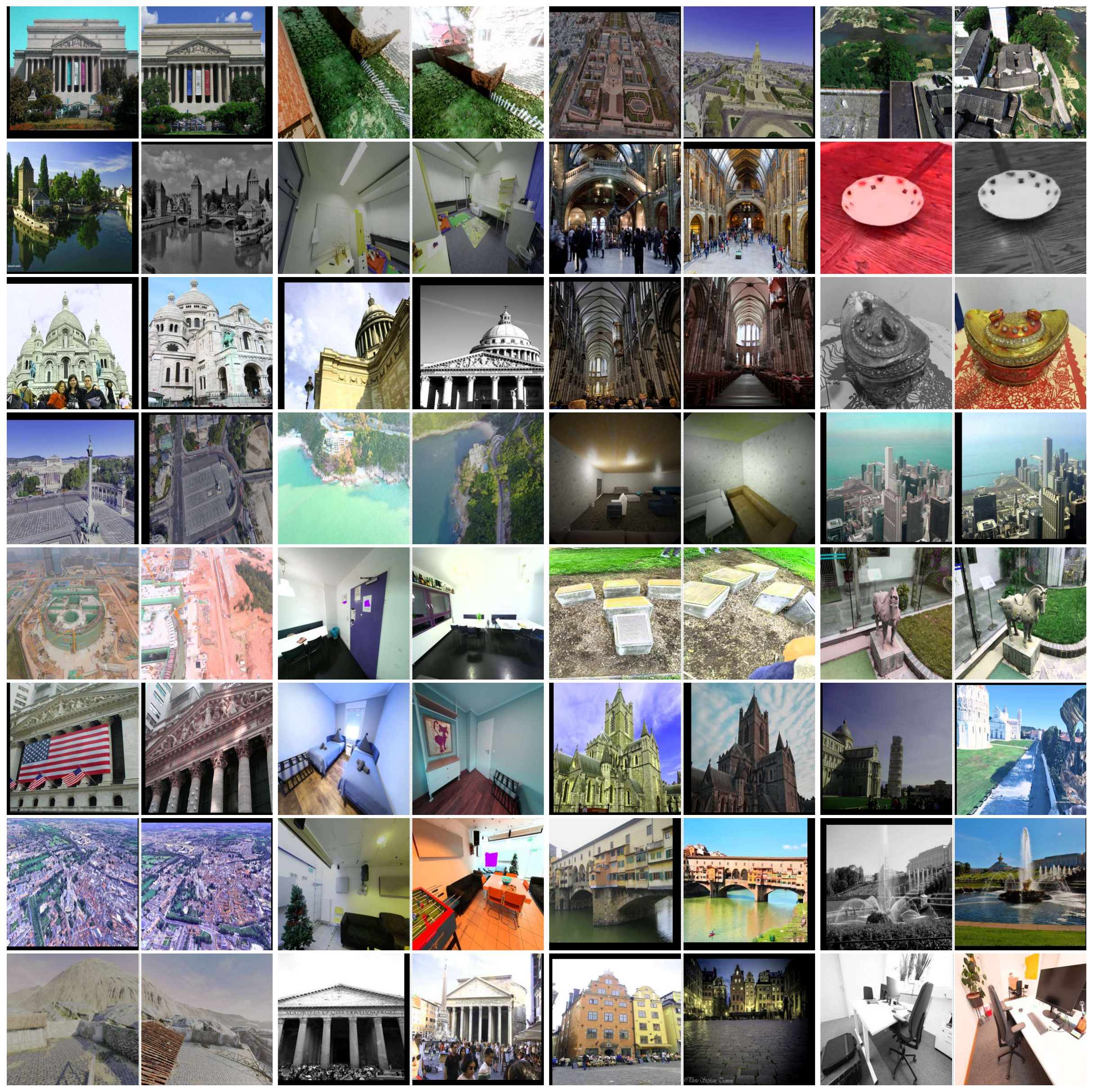}

    \caption{\textbf{Visualization of training batch.} We visualize a random training batch of 32 image pairs to highlight the diversity in our training data.}
    \label{fig:training-batch}
\end{figure}

To better understand our training data mix we randomly sample a batch of 32 image pairs and plot them in \cref{fig:training-batch}.

\end{document}